\documentclass{article}
\usepackage{iclr2025_conference,times}

\usepackage{amsmath,amsfonts,bm}

\def\eqref#1{equation~\ref{#1}}

\def\1{\bm{1}}

\def\ra{{\textnormal{a}}}

\def\re{{\textnormal{e}}}

\def\rr{{\textnormal{r}}}

\def\rva{{\mathbf{a}}}

\def\rvy{{\mathbf{y}}}
\def\rvz{{\mathbf{z}}}

\def\rmZ{{\mathbf{Z}}}

\def\va{{\bm{a}}}

\def\ve{{\bm{e}}}

\def\vo{{\bm{o}}}

\def\vr{{\bm{r}}}

\def\vy{{\bm{y}}}
\def\vz{{\bm{z}}}

\def\mA{{\bm{A}}}

\def\mI{{\bm{I}}}

\def\mO{{\bm{O}}}

\def\mY{{\bm{Y}}}

\DeclareMathAlphabet{\mathsfit}{\encodingdefault}{\sfdefault}{m}{sl}
\SetMathAlphabet{\mathsfit}{bold}{\encodingdefault}{\sfdefault}{bx}{n}

\newcommand{\E}{\mathbb{E}}

\newcommand{\R}{\mathbb{R}}

\newcommand{\sigmoid}{\sigma}

\newcommand{\Cov}{\mathrm{Cov}}

\usepackage{hyperref}
\usepackage{url}
\usepackage[utf8]{inputenc} 
\usepackage[T1]{fontenc}    
\usepackage{hyperref}       
\usepackage{url}
\usepackage{booktabs}       
\usepackage{amsfonts}       
\usepackage{nicefrac}       
\usepackage{microtype}      
\usepackage{xcolor}         

\title{Simple,\;Good,\;Fast:\;Self-Supervised\;World Models Free of Baggage}

\author{Jan Robine\rlap{,}\raisebox{0.8ex}{\normalfont\scriptsize 1,2} Marc H\"oftmann\raisebox{0.8ex}{\normalfont\scriptsize 1,2} \& Stefan Harmeling\raisebox{0.8ex}{\normalfont\scriptsize 1,2} \\ 
\textsuperscript{1}TU Dortmund, \textsuperscript{2}Lamarr Institute for Machine Learning and Artificial Intelligence \\
\texttt{\{jan.robine,marc.hoeftmann,stefan.harmeling\}@tu-dortmund.de}
}

\usepackage[textsize=tiny]{todonotes}

\usepackage{float}

\usepackage[algo2e,ruled,noend]{algorithm2e}  

\usepackage[OT1]{fontenc} 
\usepackage{enumitem}
\usepackage{tabularx}
\usepackage{siunitx}
\usepackage{relsize}
\usepackage{subcaption}
\usepackage{wrapfig}
\makeatletter
\renewcommand*{\@algocf@post@ruled}{\kern\interspacealgoruled\hrule width\algocf@ruledwidth height\algoheightrule\relax\vspace{-0.1in}}
\makeatother

\newcommand{\loss}{\mathcal{L}}
\newcommand{\VC}{\text{VC}}
\newcommand{\fenc}{f_\theta}
\newcommand{\fproj}{g_\theta}
\newcommand{\fpred}{h_\theta}
\newcommand{\ptra}{p_\theta}
\newcommand{\prew}{p_\theta}
\newcommand{\pend}{p_\theta}
\newcommand{\ppi}{\pi_\phi}
\newcommand{\fvf}{v_\phi}
\newcommand{\cond}{\:|\:}
\newcommand{\cdim}{C}
\newcommand{\hdim}{H}
\newcommand{\wdim}{W}

\newcommand{\ydim}{d}
\newcommand{\zdim}{D}

\DeclareMathOperator{\sg}{sg}
\newcommand{\xmark}{\text{\sffamily x}}  

\newcommand{\methodname}{SGF}

\usepackage[capitalize,noabbrev]{cleveref}

\newcolumntype{Y}{>{\centering\arraybackslash}X}
\newcolumntype{Z}{>{\raggedleft\arraybackslash}X}

\usepackage{tikz}
\usetikzlibrary{calc,positioning,quotes,fit}

\pgfdeclarelayer{back}
\pgfsetlayers{back,main}

\makeatletter
\pgfkeys{%
  /tikz/on layer/.code={
    \def\tikz@path@do@at@end{\endpgfonlayer\endgroup\tikz@path@do@at@end}%
    \pgfonlayer{#1}\begingroup%
  }%
}

\definecolor{niceblue}{HTML}{BBDEFB}
\definecolor{nicedarkblue}{HTML}{42A5F5}
\definecolor{nicegray}{HTML}{BDBDBD}
\definecolor{nicelightgray}{HTML}{E7E7E7}
\definecolor{nicedarkgray}{HTML}{757575}
\definecolor{nicegreen}{HTML}{C8E6C9}
\definecolor{nicedarkgreen}{HTML}{4CAF50}
\definecolor{nicepurple}{HTML}{E8CCEC}
\definecolor{nicedarkpurple}{HTML}{BA68C8}
\definecolor{nicered}{HTML}{F9BABE}
\definecolor{nicedarkred}{HTML}{E57373}
\definecolor{niceyellow}{HTML}{FFF9C4}
\definecolor{nicedarkyellow}{HTML}{FFEE58}

\tikzset{gray/.style={fill=nicegray,draw=nicedarkgray}}
\tikzset{blue/.style={fill=niceblue,draw=nicedarkblue}}
\tikzset{green/.style={fill=nicegreen,draw=nicedarkgreen}}
\tikzset{purple/.style={fill=nicepurple,draw=nicedarkpurple}}
\tikzset{red/.style={fill=nicered,draw=nicedarkred}}
\tikzset{white/.style={fill=white,draw=nicegray}}
\tikzset{yellow/.style={fill=niceyellow,draw=nicedarkyellow}}

\tikzset{var/.style={circle,line width=0.5mm,minimum size=8mm,inner sep=0mm,outer sep=0mm,text height=1.5mm,text depth=0mm}}
\tikzset{box/.style={rectangle,line width=0.5mm,rounded corners,minimum size=7mm,outer sep=0mm,text height=2mm,text depth=0mm,inner xsep=2mm}}

\tikzset{every label/.style={font=\footnotesize}}
\tikzset{every edge quotes/.style={auto,font=\footnotesize}}
\tikzset{nicearrow/.style={>={Triangle[length=2mm,width=1.75mm]},line width=0.4mm}}
\tikzset{func/.style={->,nicearrow}}
\tikzset{sim/.style={dashed,line width=0.4mm}}

\iclrfinalcopy

\begin{document}

\maketitle

\begin{abstract}
  What are the essential components of world models?
  How far do we get with world models that are not employing RNNs, transformers,
  discrete representations, and image reconstructions?
  This paper introduces \methodname{}, a Simple, Good, and Fast world model that
  uses self-supervised representation learning, captures short-time dependencies
  through frame and action stacking, and enhances robustness against model errors
  through data augmentation. We extensively discuss \methodname{}'s connections
  to established world models, evaluate the building blocks in ablation studies,
  and demonstrate good performance through quantitative comparisons on the Atari
  100k benchmark. The code is available at \url{https://github.com/jrobine/sgf}.
\end{abstract}

\section{Introduction}

Deep reinforcement learning has demonstrated remarkable success in solving
challenging decision-making problems
\citep{dqn,ppo,a3c,rainbow,agent57,muzero,meme,dreamerv3}. Despite these
achievements, the primary challenge remains sample efficiency, i.e., the amount
of data required to learn effective behaviors. Recent works have addressed this
challenge by improving architectures and hyperparameters \citep{der,bbf},
pretraining and fine-tuning \citep{spi}, applying data augmentation
\citep{drq,rad}, incorporating ideas from self-supervised representation
learning \citep{curl,spr,spi,bbf}, or learning a model of the environment
\citep{simple,efficient-zero,twm,iris,dreamerv1,dreamerv2,dreamerv3}.

\begin{wrapfigure}{r}{0.5\textwidth}
  \centering
  \vspace*{-3mm}
  \includegraphics[width=0.9\linewidth]{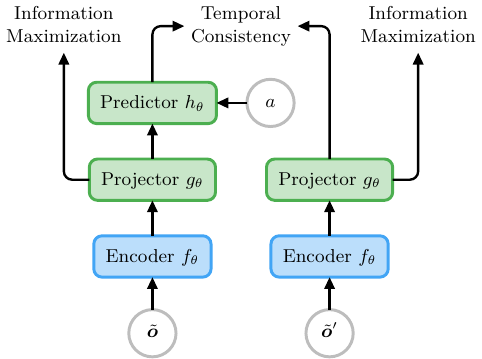}
  \vspace*{-1mm}
  \caption{Our world model learns representations that are both temporally
    consistent and maximize the information about the observations.}
  \label{fig:overview}
  \vspace*{-2mm}
\end{wrapfigure}
Several approaches have been proposed in the literature to leverage a model of
the environment. Improving the representations for the model-free agent can be
achieved by learning a (latent) transition model \citep{slac,spr}, a reward
model, or both \citep{deepmdp,dbc}, where the model serves as an auxiliary task.
Aside from that, a \emph{world model}, i.e., a deep generative model of the
environment, can be learned. World models find applications in learning in
imagination, where a model-free algorithm is applied to sequences generated by
the world model
\citep{dyna,world-models,simple,dreamerv1,dreamerv2,dreamerv3,iris,twm}, and in
decision-time planning, where the model is used for lookahead search during
action selection \citep{e2c,rce,pets,planet}. Another line of work performs
decision-time planning without a full world model, relying on \emph{value
equivalence}. In this paradigm, trajectories of the model are required to yield
the same cumulative rewards as those in the real environment, regardless of
whether the produced hidden states correspond to any real environment states or
not \citep{vi-net,predictron,vp-net,muzero,efficient-zero,td-mpc}.

In computer vision, self-supervised learning of image representations has made
significant progress in recent years
\citep{simclr,moco,byol,swav,simsiam,barlow,dino,vicreg}. Many approaches are
based on Siamese neural networks \citep{siamese} and can be categorized into
contrastive and non-contrastive methods. Contrastive methods
\citep{simclr,moco} aim to learn representations that are similar for different
views (e.g. image augmentations) of the same image but dissimilar for different
images to prevent the collapse of representations. Non-contrastive methods do
not rely on negative samples. Instead, they prevent representation collapse by
the design of the architecture \citep{byol,simsiam} or by regularization of the
representations \citep{barlow,vicreg}.

\vspace{-1pt}  
\paragraph{Contributions:}
In this work, we explore simplifying world models while maintaining good
performance. Many world models rely on RNNs or transformers to capture long-term
dependencies, introducing computational complexity and instability. We focus on
problems where short-term dependencies might be sufficient, \emph{avoiding
sequence models} and instead leveraging self-supervised learning, data
augmentation, and stacking.
%
%
%
Our contributions are as follows:
\begin{itemize}[topsep=0pt]
  \item We present \methodname{}, a world model based on a representation learning
        framework inspired by VICReg \citep{vicreg}. While restricting ourselves to
        simple world models (w/o RNNs and transformers) and choosing simple
        ingredients, we still have the essential properties of effective world models:
        \emph{maximum information} and \emph{temporal consistency}
        (\cref{sec:properties}).
  \item We reduce the complexity of world models in terms of both methodology and
        implementation: \methodname{} does not require image reconstructions or
        discretization of representations. Besides avoiding sequence models, such as
        recurrent neural networks or transformers, we also do not use probabilistic
        predictions for deterministic environments. Instead, we employ simple
        techniques from model-free reinforcement learning, in particular data
        augmentation and stacking, which already have been successfully used before
        (\cref{sec:ingredients}).
  \item We conduct several ablations and thoroughly discuss the similarities and
        differences between \methodname{} and other world models that learn in
        imagination (\cref{sec:ablations,sec:methodology}).
  \item We demonstrate that our design choices lead to shorter training times compared
        to other world models, while achieving good performance on the Atari 100k
        benchmark (\cref{sec:results}).
\end{itemize}


\section{Simple, Good, and Fast World Models}
\label{sec:sgf}

While being increasingly powerful, model-based approaches have simultaneously
grown in complexity and consist of more and more components that need to be
adjusted to each other \citep{dreamerv1, dreamerv2, dreamerv3, twm, iris}.
After introducing our notation, we aim to distill minimal ingredients and
objectives for world models that are easy to implement and computationally
efficient.

\subsection{Preliminaries and Notation}

We formalize the environment in terms of a partially observable Markov decision
process (POMDP) with discrete time steps, rewards ${r \in \R}$, image
observations ${\vo \in \R^{\cdim \times \hdim \times \wdim}}$, and actions ${a
      \in \mathcal{A}}$, which are either discrete or continuous. Transitions within
the environment are described by a tuple ${(\vo, \va, \vo', r, e)}$, where
${\vo}$ is the current observation, $\vo'$ is the next observation, and ${e \in
      \{0, 1\}}$ indicates terminal states.

In its simplest form, a world model is a generative model of the dynamics
${p(\vo', r, e \cond \vo, \va)}$ of a POMDP. Given a policy ${\va = \pi(\vo)}$,
iterative sampling from the world model generates trajectories without further
real environment interactions. These trajectories can be used for learning
behaviors in imagination \citep{world-models,dreamerv1}, e.g., via model-free
RL.

To increase efficiency, world models should operate in a low-dimensional
representation space (commonly known as latent space) as opposed to the
high-dimensional observation space \citep{world-models}. For this we need two
components: a \emph{representation model} that maps image observations $\vo$
onto representations $\vy$, and a \emph{dynamics model} that predicts the
latent dynamics ${p(\vy', r, e \cond \vy, \va)}$. Usually, the policy also
operates in the low-dimensional space, i.e., ${\va = \pi(\vy)}$, enabling
behavior learning with high computational efficiency \citep{world-models}.

A common assumption is the conditional independence of ${r}$ and ${e}$ given
${\vy}$, ${\va}$, and ${\vy'}$, which is also employed by previous world models
\citep[e.g.,][]{iris,dreamerv3}. This leads to the following factorization of
the latent dynamics
\begin{equation}
  \label{eq:dynamics}
  p(\vy', r, e \cond \vy, \va) = p(\vy' \cond \vy, \va) \, p(r \cond \vy, \va, \vy') \, p(e \cond \vy,
  \va, \vy'),
\end{equation}
which consists of three conditional distributions: the \emph{transition
  distribution} ${p(\vy' \cond \vy, \va)}$, which is only conditioned on $\vy$ and
$\va$, the \emph{reward distribution} ${p(r \cond \vy, \va, \vy')}$ and the
\emph{terminal distribution} ${p(e \cond \vy, \va, \vy')}$, both further
conditioned on the next representation $\vy'$. This is a natural choice for many
POMDPs, where most of the complexity of the dynamics is captured by the
transition distribution. This allows for learning three separate models rather
than modeling the joint distribution.




\subsection{Ingredients Leading to Simplicity}
\label{sec:ingredients}


\paragraph{Stacking instead of memory.}
Model-free methods often assume that the observations of an POMDP approximately
satisfy the Markov property. This means that the next observation ${\vo'}$ (and
consequently ${\vy'}$) is independent of the preceding history of transitions
given the current observation $\vo$ and action $\va$. However, previous works on
world models consider \emph{non}-Markovian observations, which might exhibit
long-term dependencies. This is approached by adding a notion of \emph{memory}
to the dynamics model, which can be realized by introducing recurrent states
\citep[RNN-based,][]{world-models, dreamerv1} or by directly conditioning on the
history of transitions \citep[attention-based][]{twm,iris}. However, the memory
is typically a big computational burden and makes the model more complicated. We
are asking whether we can omit the memory to obtain a much faster world model.

To capture short-time dependencies with minimal computational overhead, we
suggest to simply use \emph{frame and action stacking}. Stacking the $m$ most
recent \emph{frames} alleviates the problem of partial observability, e.g., by
capturing the velocity of objects in the scene. This is a well-known
preprocessing technique in model-free reinforcement learning \citep{dqn}, and
has already been applied to world models by \citet{twm}. Additionally, stacking
the most $m$ recent \emph{actions} can be beneficial, considering potential
delays in the effects of actions.
In \cref{sec:ablations}, we show significant improvements of our world model
through action stacking while being computationally cheap.


\paragraph{Augmentations instead of stochasticity.}
In deterministic POMDPs, executing a specific action in a specific state
consistently yields the same outcomes ${\vo'}$, ${r}$, and ${e}$. However,
prior world models are stochastic even in deterministic POMDPs (see
\cref{sec:methodology}). \citet{world-models} argue that, due to model errors,
behaviors learned in imagination may perform poorly in the real environment.
They propose that stochastic predictions can reduce the exploitability of an
imperfect world model.

Similarly, we introduce stochasticity through \emph{data augmentation}, as
demonstrated in previous model-free algorithms \citep{drq,rad}. We investigate
whether this helps to improve the robustness of our world model against model
errors. In \cref{sec:ablations}, we demonstrate that data augmentation
significantly improves the performance of our world model.



\subsection{Essential Properties of Representations}
\label{sec:properties}

Building meaningful representations of observations is crucial for dynamics
modeling and behavior learning. In this work, we argue that representations
should possess two key properties: (1) the information about the observations
should be maximized, (2) they should be temporally consistent, i.e.,
representations of two successive observations should be similar. We describe
these properties in more detail below. Both properties are already present in
other world models (see \cref{sec:methodology}), however, in this work we try
to implement them in a most simple and efficient manner.

\paragraph{Maximizing information.}
In latent world models, the representations are used for downstream behavior
learning, so any information not encoded in these representations is not
accessible to the agent. Therefore, extracting maximum information from
observations is necessary for learning optimal behaviors in latent space. This
is often realized by reconstructing the input observations (i.e.,
autoencoder-style), however, this can also be realized by self-supervised
objectives, as we show in \cref{sec:building-blocks}.

\paragraph{Temporal consistency.}
Temporal consistency can be motivated by \textit{predictive coding}, where the
future or missing information is predicted. Predictive coding has been applied
in information theory for data compression \citep{predictive-coding}, and more
recently in representation learning \citep{cpc,cpcv2}. Also, the neuroscience
literature suggests that the human brain learns internal representations of
incoming sensory signals by minimizing prediction errors subject to particular
constraints on the representations, in spatial and temporal domains
\citep{predictive-visual,predictive-retina,predictive-review}. Furthermore,
\citet{perceptual-straightening} proposed the \emph{temporal straightening
  hypothesis} which suggests that inside the human brain visual inputs are
transformed to follow straighter temporal trajectories, in order to make the
stream of visual inputs more predictable.
These advantages of \emph{temporal consistency} for biological agents translate
to advantages for agents in reinforcement learning that are based on latent
world models:
\begin{itemize}[topsep=0pt]
  \item \textit{Simpler dynamics prediction:} as successive observations are
        close in representation space, the dynamics model often only needs to
        predict minor changes and the danger of sudden jumps in representation
        space is reduced.
  \item \textit{Improved behavior learning:} both, the policy and value function
        benefit from temporally similar representations. This concept can be
        loosely connected to \emph{bisimulation metrics}, where ``behaviorally
        similar'' states are grouped together \citep{dbc}. Further details are
        available in \cref{sec:rec-free}.
\end{itemize}

\section{Building Blocks for \methodname{}}
\label{sec:building-blocks}

Having identified core ingredients, we now describe the construction of a
simple, fast, and good world model. We will begin by outlining the
representation model of \methodname{}, followed by an description of the
dynamics model, implementational details, and our evaluation protocol. For
representation learning we draw inspiration from VICReg \citep{vicreg}. Further
connections to other self-supervised methods are discussed in \cref{sec:ssl}.

\subsection{Learning a World Model} \label{sec:meth}

\paragraph{Representation learning.}
Given a POMDP transition ${(\vo, \va, \vo', r, e)}$, we apply random
transformations $t, t' \sim \mathcal{T}$ from a set $\mathcal{T}$ of image
augmentations to obtain augmented observations ${\tilde{\vo} = t(\vo)}$ and
${\tilde{\vo}' = t'(\vo')}$. An encoder $\fenc$ computes representations
${\tilde{\vy} = \fenc(\tilde{\vo})}$ and ${\tilde{\vy}' = \fenc(\tilde{\vo}')}$
with ${\tilde{\vy}, \tilde{\vy}' \in \R^\ydim}$. A projector network $\fproj$
computes embeddings ${\tilde{\vz} = \fproj(\tilde{\vy})}$ and ${\tilde{\vz}' =
      \fproj(\tilde{\vy}')}$ with ${\tilde{\vz}, \tilde{\vz}' \in \R^\zdim}$. An
action-conditioned predictor network $\fpred$ predicts the next embedding
${\hat{\vz}' = \fpred(\tilde{\vz}, a)}$. An illustration can be seen in
\cref{fig:overview}.

To achieve temporal consistency, we minimize the mean squared error between
$\hat{\vz}'$ and $\tilde{\vz}'$. To maximize the information content, the
embeddings are regularized using the variance and covariance regularization
terms proposed by \citet{vicreg}. The total representation loss is summarized
by
\begin{equation}
  \label{eq:repr-loss}
  \loss_\text{Repr.}(\theta) = \E_\tau \Big[ \,\underbrace{ \tfrac{\eta}{\zdim} \Vert \fpred(\tilde{\rvz}, \rva) - \tilde{\rvz}' \Vert_2^2}_\text{Temporal Consistency} {}+{} \underbrace{\vphantom{\tfrac{1}{d}} \VC(\tilde{\rmZ}) + \VC(\tilde{\rmZ}')}_\text{Information Maximization}\, \Big],
\end{equation}
where $\tau$ is a batch of transitions from a replay buffer, $\tilde{\rmZ}$ and
$\tilde{\rmZ}'$ are batches of embeddings, ${\eta > 0}$ controls the strength
of the consistency loss, and $\VC$ (variance and covariance) is defined as
\begin{equation}
  \label{eq:vc-loss}
  \VC(\rmZ) = \frac{1}{\zdim} \sum_{j=1}^\zdim \bigg[ \,\underbrace{\vphantom{\sum\nolimits_{k \neq j}} \rho \max\!\left( 0, 1 - \sqrt{\smash[b]{\Cov(\rmZ)_{j,j}} + \varepsilon} \right)}_\text{Variance Regularization} + \underbrace{\nu \sum\nolimits_{k \neq j} \Cov(\rmZ)_{j,k}^2}_\text{Covariance Regularization}\, \bigg],
\end{equation}
where $\zdim$ is the dimensionality of the embeddings, ${\rho, \nu > 0}$ control
the strength of variance and covariance regularization terms, respectively, and
${\varepsilon = \num{1e-4}}$ prevents numerical instabilities. The goal of the VC terms
is to maximize information content and to prevent representation collapse.
\textbf{V}ariance regularization keeps the standard deviation of each embedding
feature across the batch above $1$ using a hinge loss. \textbf{C}ovariance
regularization decorrelates the embedding features by attracting their
covariances towards zero \citep{vicreg}.

\paragraph{Dynamics learning.}

We build a simple dynamics model and rely on the capabilities of temporally
consistent representations. Based on the dynamics factorization, we learn a
transition distribution ${\ptra(\vy' \cond \vy, \va)}$, a reward distribution
${\prew(r \cond \vy, \va, \vy')}$, and a terminal distribution ${\pend(e \cond
      \vy, \va, \vy')}$. In this work, we focus on deterministic prediction, i.e., we
simply calculate the means for transitions and rewards, and the mode for
terminals (more details on this in \cref{sec:impl-details}). Maximum likelihood
estimation leads to the total dynamics loss
\begin{equation}
  \label{eq:dyn-loss}
  \loss_\text{Dyn.}(\theta) = \E_\tau \Big[ - \underbrace{ \log \ptra(\sg(\rvy') \cond \sg(\rvy), \rva) }_\text{Transition Distribution} -\underbrace{\vphantom{\tfrac{1}{d}} \log \prew(\rr \cond \tilde{\rvy}, \ra, \tilde{\rvy}')}_\text{Reward Distribution} {}-{} \underbrace{\vphantom{\tfrac{1}{d}} \log \pend(\re \cond \tilde{\rvy}, \ra, \tilde{\rvy}')}_\text{Terminal Distribution}\, \Big],
\end{equation}
where ${\sg(\cdot)}$ denotes the stop-gradient operator, meaning that the
representations are not influenced by the loss of the transition distribution.
This is because the transition distribution has moving targets, given that
${\vy'}$ originates from the representation model, which changes during
training. The rewards and terminals provide stable signals from the POMDP. Note
that we learn the transitions with non-augmented observations, i.e., ${\vy =
      \fenc(\vo)}$ and ${\vy' = \fenc(\vo')}$.




\subsection{Learning Behaviors in Imagination} \label{sec:imagination}

The representations $\vy$ serve as inputs to the policy ${\ppi(\va \cond
      \vy)}$. Through iterative dynamics prediction, batches of representations,
actions, rewards, and terminals are generated and used to train the policy. The
policy learns to maximize the expected return by performing approximate
gradient ascent with the policy gradient \citep{policy-gradient}. To reduce the
variance of the gradient estimates, we employ a learned value function
${\fvf(\vy)}$ as a baseline, resulting in an advantage actor-critic approach
\citep{a3c}; the details are explained in \cref{sec:impl-details}. The
pseudocode outlining our world model and policy training procedure is presented
in \cref{algo:main}.



\subsection{Evaluation Protocol}

We evaluate our world model on the Atari 100k benchmark, which was first
proposed by \citet{simple} and has been used to evaluate many sample-efficient
reinforcement learning methods \citep{curl,drq,spr,bbf,iris,dreamerv3}. It
includes a subset of 26 Atari games from the Arcade Learning Environment
\citep{ale} and is limited to 400k environment steps, which amounts to 100k
steps after frame skipping or 2 hours of human gameplay. Note that all games
are deterministic \citep{revisiting}.

We perform 10 runs per game and for each run we compute the average score over
100 episodes at the end of training. We follow \citet{iris} by selecting a
random action with $1\%$ probability inside the environment and using a
sampling temperature of $0.5$ for the policy during evaluation. We also adapt
their special handling of Freeway and use a sampling temperature of $0.01$ for
the policy.

\section{Empirical Study} \label{sec:emp-study}

Behavior learning depends entirely on the quality of the world model. To assess
our world model, we first analyze it qualitatively and then show how getting
rid of individual design choices results in a degradation of performance.
Further analysis and ablations are presented in
\cref{sec:add-analysis,sec:add-ablations}.

\begin{figure}[t]
  \centering
  \begin{minipage}[b]{0.48\textwidth}
    \centering
    \begin{tikzpicture}[
        image/.style={inner sep=0pt, outer sep=0pt},
        node distance=2mm and 1mm
      ]
      \node [image] (frame1-1) {\includegraphics[width=1.56cm]{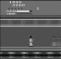}};
      \node [image, right=of frame1-1] (frame1-2) {\includegraphics[width=1.56cm]{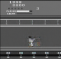}};
      \node [image, right=of frame1-2] (frame1-3) {\includegraphics[width=1.56cm]{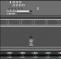}};
      \node [image, right=of frame1-3] (frame1-4) {\includegraphics[width=1.56cm]{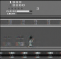}};

      \node [image, below=of frame1-1] (frame2-1) {\includegraphics[width=1.56cm]{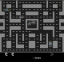}};
      \node [image, right=of frame2-1] (frame2-2) {\includegraphics[width=1.56cm]{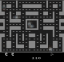}};
      \node [image, right=of frame2-2] (frame2-3) {\includegraphics[width=1.56cm]{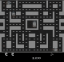}};
      \node [image, right=of frame2-3] (frame2-4) {\includegraphics[width=1.56cm]{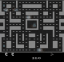}};

      \node [image, below=of frame2-1] (frame3-1) {\includegraphics[width=1.56cm]{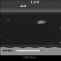}};
      \node [image, right=of frame3-1] (frame3-2) {\includegraphics[width=1.56cm]{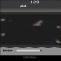}};
      \node [image, right=of frame3-2] (frame3-3) {\includegraphics[width=1.56cm]{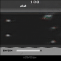}};
      \node [image, right=of frame3-3] (frame3-4) {\includegraphics[width=1.56cm]{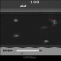}};
    \end{tikzpicture}
    \vspace{1.5mm}
    \captionof{figure}{Illustration of imagined sequences of length 30. Each frame in the
      frame stack is converted to grayscale, and pixel changes are visualized in
      red, green, and blue.
      From top to bottom: Kung Fu Master, Ms Pacman, and Seaquest.}
    \label{fig:dreams}
  \end{minipage}
  \hfill
  \begin{minipage}[b]{0.48\textwidth}
    \centering
    \includegraphics[width=\linewidth,trim={2.3cm 0.85cm 0.95cm 1.05cm},clip]{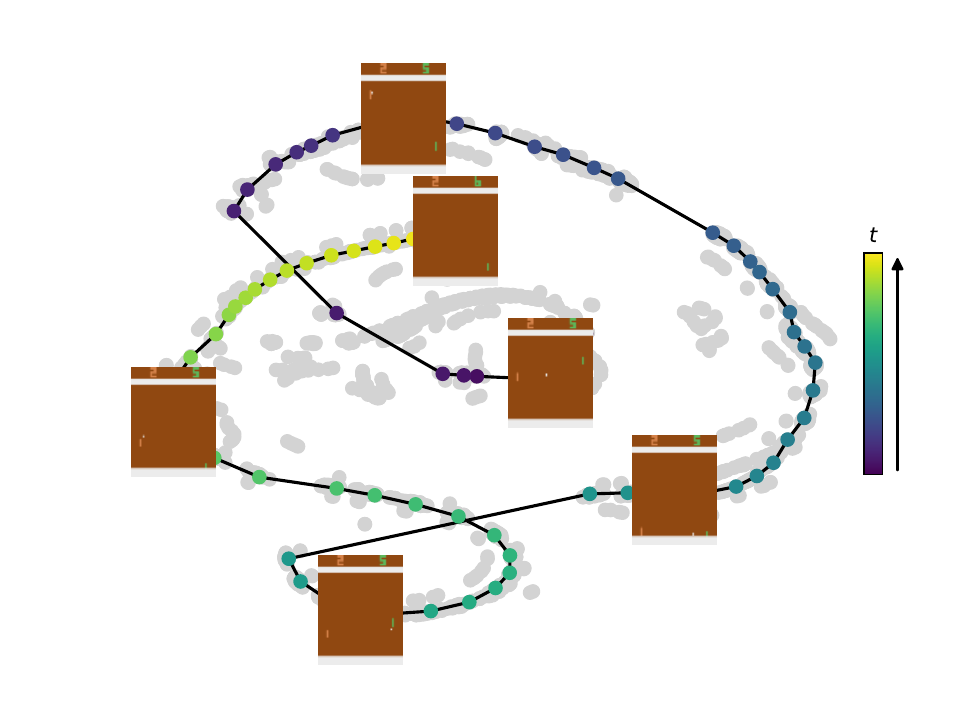}
    \captionof{figure}{Two-dimensional t-SNE embeddings of the learned representations
      obtained by playing an episode in Pong. We highlight a subsequence of the
      episode, which starts with a new ball and stops after a point is scored.}
    \label{fig:tsne}
  \end{minipage}
\end{figure}


\subsection{Inspecting the World Model}
\label{sec:analysis}

To assess whether the learned representations contain relevant information and
do not collapse, we train a separate decoder for analysis without affecting the
world model's gradients. This allows us to visually interpret the imagined
sequences of the world model. In \cref{fig:dreams} we depict three exemplary
sequences, demonstrating that the learned representations contain useful
information without relying on image reconstructions. We can also see that the
dynamics model can work for long sequences (30 time steps) without accumulating
notable model errors, although being only a feedforward model.

Furthermore, we study the temporal consistency of the learned representations.
For that, we generate an episode in Pong following a policy trained with our
approach. We then encode the observations and compute two-dimensional t-SNE
embeddings \citep{tsne} of the representations. The result can be seen in
\cref{fig:tsne}, and it suggests that subsequent observations have similar
representations and temporally consistency is successfully employed. In
\cref{fig:analyze-consistency} we compare the learned embeddings when disabling
temporal consistency.

\subsection{Ablating the World Model}
\label{sec:ablations}

\begin{figure}[t]
  \centering
  \begin{minipage}[b]{0.48\textwidth}
    \centering
    \includegraphics[width=\linewidth]{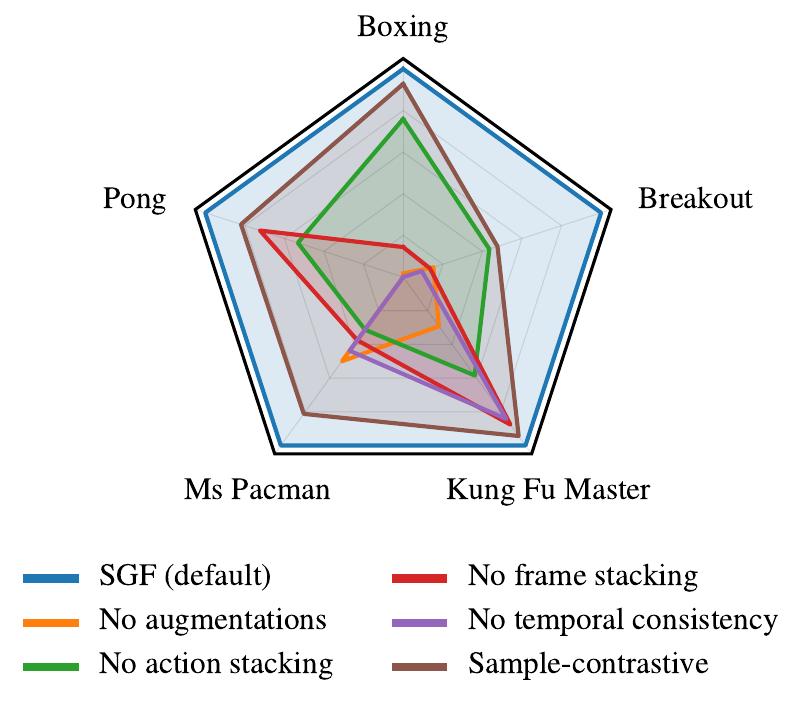}
    \captionof{figure}{Ablations of \methodname{} in five games. Human
      normalized scores, normalized with the maximum value achieved per game.}
    \label{fig:ablations}
  \end{minipage}
  \hfill
  \begin{minipage}[b]{0.48\textwidth}
    \centering
    \includegraphics[width=\linewidth,trim={5mm 5mm 5mm 5mm},clip]{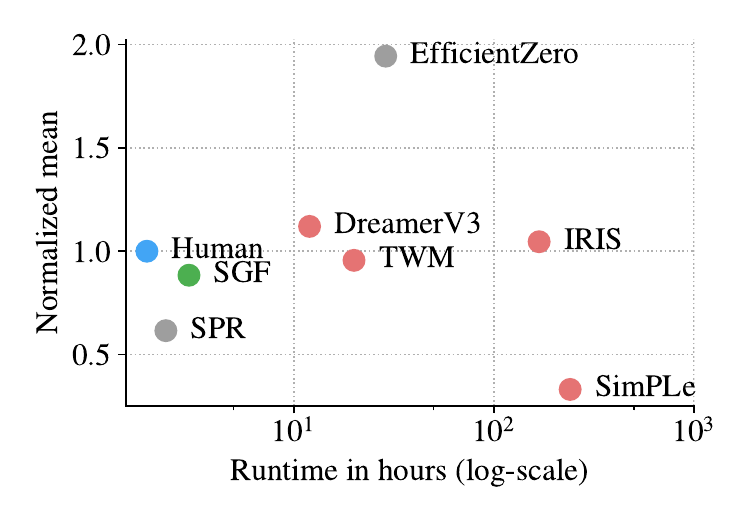}
    \captionof{figure}{Score and runtime comparison in the Atari 100k benchmark.
      SPR is model-free, EfficientZero performs lookahead.}
    \label{fig:runtimes}
  \end{minipage}
\end{figure}

We show empirically how effective the components presented in \cref{sec:sgf}
are by performing five ablations. Each ablation is assessed by the performance
on five Atari games. The results are illustrated in \cref{fig:ablations}.
Numerical results can be found in \cref{tab:ablations}. We observe that data
augmentation, action stacking, frame stacking, and temporal consistency are
crucial:
\begin{enumerate}[topsep=0pt]
  \item \emph{No augmentations}: omitting image augmentations leads to poor
        performance in all games.
  \item \emph{No action stacking}: stacking only the frames but not the actions
        decreases the overall performance for all five games.
  \item \emph{No frame stacking}: stacking only the actions but not the frames
        decreases the performance of all games, with complete failures in Boxing
        and Breakout. However, Kung Fu Master suffers only a small degradation,
        possibly because all enemies face the direction they are heading, so
        their velocity is identifiable from a single frame.
  \item \emph{No temporal consistency}: setting the coefficient $\eta$ to zero
        leads to a significant decrease in performance for most games except
        Kung Fu Master.
  \item \emph{Sample-contrastive}: \citet{duality} show that VICReg can be also
        seen as a \emph{dimension-contrastive} method, as opposed to
        \emph{sample-contrastive} methods such as SimCLR \citep{simclr}. They
        also show that VICReg can be converted to a sample-contrastive method by
        transposing the embedding matrix $\rmZ$ in \cref{eq:vc-loss}. Making
        VICReg sample-contrastive significantly worsens the performance in
        Breakout, but less so in the other games.
\end{enumerate}

\section{Comparisons}

In this section we compare our method with previous world models regarding the
results and the methodology. In \cref{sec:add-relations} we discuss the
relations to other methods.

\subsection{Result Comparison}
\label{sec:results}

We compare our method with five baselines: the model-free algorithm SPR
\citep{spr}, with updated scores from \citet{rl-eval}, and the model-based
methods EfficientZero \citep{efficient-zero}, IRIS \citep{iris}, and DreamerV3
\citep{dreamerv3}. The mean metric across all games is calculated using human
normalized scores \citep{dqn}.

In terms of performance, our simple world model achieves good scores with
significantly faster training. \cref{fig:runtimes} shows the scores in relation
to runtime. Detailed scores can be found in \cref{tab:results} and numerical
runtimes can be found in \cref{tab:runtimes}. Training \methodname{} takes 1.5
hours on a single NVIDIA A100 GPU. Obtaining precise training times for other
methods is challenging, as they depend on the GPU. Following \citet{dreamerv3},
we approximate runtimes for an NVIDIA V100 GPU, assuming NVIDIA P100 GPUs are
twice as slow and NVIDIA A100 GPUs are twice as fast. Notably, \methodname{}'s
runtime is four times shorter than the runtime of DreamerV3, despite both
having the same number of imagination steps ($1.5$ billion). In
\cref{tab:time-details} we provide a breakdown of the runtime for certain
components of our method.


\subsection{World Model Comparison}
\label{sec:methodology}

We also compare the methodologies of \methodname{} and state-of-the-art world
models used for learning in imagination: SimPLe \citep{simple}, Dreamer
\citep{dreamerv1,dreamerv2,dreamerv3}, IRIS \citep{iris}, TWM \citep{twm}, and
the world model developed by \citet{world-models}, referred to as HS. In the
Dreamer line of work, our focus is on DreamerV2 and DreamerV3, given their
performance improvements over the initial version, notably achieved through the
discretization of representations. A summarized comparison of the main
differences is provided in \cref{tab:meth-comp} (we highlight the components in
the text).



\paragraph{Temporal consistency.}

\begin{figure}
  \centering
  \begin{subfigure}[t]{0.49\linewidth}
    \centering
    \includegraphics{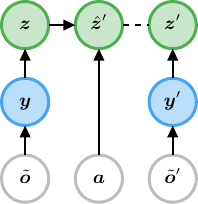}
    \caption{\methodname{} (ours)}
  \end{subfigure}
  \hfill
  \begin{subfigure}[t]{0.49\linewidth}
    \centering
    \includegraphics{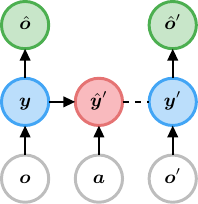}
    \caption{Dreamer}
  \end{subfigure}

  \caption{High-level illustration of how \methodname{} and Dreamer ensure temporal
    consistency. The dashed lines denote that two variables are attracted
    towards each other by a loss function. White nodes indicate inputs, blue
    nodes indicate representations, and green nodes indicate variables used for
    representation learning. The red node indicates that Dreamer depends on the
    dynamics model for representation learning. We omit Dreamer's recurrent
    states for simplicity.}
  \label{fig:dreamer-comp}
  \vskip -0.2in
\end{figure}

We enforce successive representations to be similar by the choice of our
objective (\emph{Consistency}). IRIS and HS have no explicit concept of
temporal consistency. Dreamer and TWM also seek temporal consistency and
attract the representations slightly towards the outputs of the transition
predictor from the previous time step. The transition predictor can be
interpreted as a time-dependent prior for a variational auto-encoder. An
illustration of this difference can be seen in \cref{fig:dreamer-comp}. Our
approach is simpler for the following reasons:
\begin{itemize}[topsep=0pt]
  \item In Dreamer, the training of the representation model and the dynamics model is
        intertwined. Correctly balancing the representation loss and the dynamics loss
        is crucial for ensuring stable training. Our representation model learns in
        isolation, simplifying hyperparameter tuning while still achieving temporal
        consistency.
  \item In Dreamer, consistency is imposed directly on the representations, whereas we
        maximize the similarity of the non-linear embeddings of the representations. In
        DreamerV3 \citep{dreamerv3} the consistency loss is clipped when it falls below
        a certain threshold, considering the similarity as sufficient (aka free bits).
        Our hypothesis is that maximizing the similarity of the embeddings offers a
        similar degree of freedom. 
\end{itemize}

\begin{table}
  \centering
  \caption{Comparison of methodology with other world models used for
    imagination. DVx denotes DreamerV2 and DreamerV3. Every
    component results in additional complexity.}
  \label{tab:meth-comp}
  \vspace{\belowcaptionskip}
  \small
  \begin{tabularx}{\linewidth}{lYYYYYY}
    \toprule
    \textbf{Component}       & \textbf{HS} & \textbf{SimPLe} & \mbox{\hspace{-3pt}\textbf{IRIS}} & \mbox{\hspace{-4pt}\textbf{TWM}} & \mbox{\hspace{-1pt}\textbf{DVx}} & \mbox{\hspace{-2pt}\textbf{\methodname{}}} \\
    \midrule
    Augmentations            &             &                 &                                   &                                  &                                  & \xmark                                     \\
    Information Maximization &             &                 &                                   &                                  &                                  & \xmark                                     \\
    Stacking                 &             &                 &                                   & \xmark                           &                                  & \xmark                                     \\
    Consistency              &             &                 &                                   & \xmark                           & \xmark                           & \xmark                                     \\
    Reconstructions          & \xmark      & \xmark          & \xmark                            & \xmark                           & \xmark                           &                                            \\
    Discretization           &             & \xmark          & \xmark                            & \xmark                           & \xmark                           &                                            \\
    Sequential Dynamics      & \xmark      &                 & \xmark                            & \xmark                           & \xmark                           &                                            \\
    Stochastic Transitions   & \xmark      & \xmark          & \xmark                            & \xmark                           & \xmark                           &                                            \\
    Pixel Transitions        &             & \xmark          &                                   &                                  &                                  &                                            \\
    Pixel Dreams             &             & \xmark          & \xmark                            &                                  &                                  &                                            \\
    Act with Memory          & \xmark      &                 & \xmark                            & \xmark                           & \xmark                           &                                            \\
    %
    \bottomrule
  \end{tabularx}
\end{table}

\paragraph{Information extraction.}
Previous world models depend on pixel-wise image reconstruction for information
extraction from observations (\emph{Reconstructions}). These auto-encoder
architectures treat all pixels equally, including less important high-frequency
details or noise. In contrast, we adopt a self-supervised objective and utilize
data augmentation (\emph{Augmentations}) to learn representations that maximize
information (\emph{Information~Maximization}) and extract relevant features.

\paragraph{Discretization.}


Most previous methods learn discrete representations (\emph{Discretization}):
SimPLe discretizes representation values into bits. DreamerV2
\citep{dreamerv2}, DreamerV3 \citep{dreamerv3}, and TWM \citep{twm} utilize
softmax normalization to obtain a stack of independent categorical
distributions. IRIS \citep{iris} converts each image observation into multiple
discrete tokens. Discretization introduces additional complexity and requires
techniques such as straight-through gradient estimation
\citep{straight-through}. We propose two primary factors for the success of
discretization in world models and explain how they are addressed in our
approach:
\begin{itemize}[topsep=0pt]
  \item Discretization significantly limits the information capacity of
        representations, potentially preventing collapse in auto-encoder architectures
        \citep{path}. Since our objective already prevents representation collapse,
        there is no need for discretization on that account.
  \item Discretization potentially facilitates dynamics prediction by shrinking and
        stabilizing the support of ${p(\vy' \cond \vy, \va)}$. However, we found that a
        simple architectural choice, specifically adding layer normalization as the
        final layer of the encoder (as mentioned in \cref{sec:impl-details}), is
        sufficient to keep the mean and variance of the representations stable. Note
        that another common approach is to normalize the outputs to lie in the interval
        $[0,1]$ \citep{muzero,spr}.
\end{itemize}

\paragraph{Dynamics modeling.}
As we assume Markovian observations, our approach utilizes a feedforward
dynamics model. Prior methods, with the exception of SimPLe, are RNN-based (HS,
DreamerV2, DreamerV3) or transformer-based (IRIS, TWM)
(\emph{Sequential~Dynamics}). SimPLe operates directly on image observations
instead of operating in a low-dimensional representation space
(\emph{Pixel~Transitions}). Additionally, our dynamics model is deterministic,
whereas previous methods are stochastic (\emph{Stochastic~Transitions}), at
least regarding transition prediction, while rewards and terminals typically
remain deterministic.

\paragraph{Behavior learning.}
Efficient world models, such as ours, usually train a policy based on the
low-dimensional representations. However, IRIS decodes the representations back
to pixels, slowing down training significantly (\emph{Pixel Dreams}). Since
SimPLe predicts the pixels directly, their policy has to operate on pixels as
well. Moreover, previous methods that use a sequence model usually equip the
policy with a memory, since the representations encapsulate the history of
transitions either via attention or through a compressed recurrent state
(\emph{Act~with~Memory}). Due to our feedforward dynamics model, our policy is
memoryless.

\paragraph{Stacking.}
Previous world models considered in this section only employ the usual
preprocessing, e.g., conversion to grayscale or downscaling the observations.
Our method applies frame and action stacking (\emph{Stacking}); TWM applies
only frame stacking.


\section{Limitations}
\label{sec:limitations}

The insights presented in our paper provide a solid foundation for new world
models that are simple, good, and fast. For this paper, we limited our method to
deterministic MDPs. To also include non-deterministic POMPDs, the transition
distribution needs to be stochastic, allowing for the prediction of multiple
possible outcomes. The predictor must also be stochastic to account for
non-deterministic information between $o$ and $o'$. Both networks would need to
make stochastic predictions, e.g., by modeling the mean and variance of
independent normal distributions or using Gaussian mixtures.

Having avoided sequence models such as RNN and transformers, we did limit
ourselves to environments with mainly short-term dependencies, which might be
the reason that we did not reach state-of-the-art performance. Future work could
replace the MLPs used for the transition, reward, and terminal distributions
with a sequence model, requiring some implementation effort and resulting in
increased computation time. Since the transition distribution is independent of
representation learning, this change would not affect the encoder. We evaluated
a preliminary version of this in Appendix E. Another more sophisticated approach
would involve using a sequence model for the predictor, which would also affect
the features extracted by the encoder. However, this would significantly
increase the complexity of the model, which we aimed to avoid.

Another current limitation of our approach is that VICReg requires the
observations to be images. In principle, VICReg could be applied to other
modalities if reasonable augmentations are available. We could also combine
\methodname{} with other self-supervised learning methods.

\section{Conclusion}
\label{sec:conclusion}

The starting point of our work are the questions: What are the essential
components of world models? How far do we get with world models that are not
employing RNNs, transformers, discrete representations, and image
reconstructions? We demonstrate that representations learned in a
self-supervised fashion using VICReg combined with a action-conditioned
predictor network and applied to stacked observations can learn latent
representations without resorting to resource-intensive sequence models.
Self-supervised learning, coupled with augmentations and frame and action
stacking, proves effective in building a good world model. Applying
\methodname{} to the Atari 100k benchmark, we attained good results with
significantly reduced training times. We advocate for future research in
model-based reinforcement learning to focus on sparingly adding new components
and analyzing their necessity under varying circumstances.




\subsubsection*{Acknowledgments}
This research has been funded/supported by the Federal Ministry of Education and
research of Germany and the state of North Rhine-Westphalia as part of the
Lamarr Institute for Machine Learning and Artificial Intelligence.



\bibliography{main}

\begin{thebibliography}{67}
\providecommand{\natexlab}[1]{#1}
\providecommand{\url}[1]{\texttt{#1}}
\expandafter\ifx\csname urlstyle\endcsname\relax
  \providecommand{\doi}[1]{doi: #1}\else
  \providecommand{\doi}{doi: \begingroup \urlstyle{rm}\Url}\fi

\bibitem[Agarwal et~al.(2021)Agarwal, Schwarzer, Castro, Courville, and
  Bellemare]{rl-eval}
Rishabh Agarwal, Max Schwarzer, Pablo~Samuel Castro, Aaron~C. Courville, and
  Marc~G. Bellemare.
\newblock Deep reinforcement learning at the edge of the statistical precipice.
\newblock In \emph{Advances in Neural Information Processing Systems 34: Annual
  Conference on Neural Information Processing Systems 2021, NeurIPS 2021,
  December 6-14, 2021, virtual}, pp.\  29304--29320, 2021.

\bibitem[Ba et~al.(2016)Ba, Kiros, and Hinton]{layer-norm}
Lei~Jimmy Ba, Jamie~Ryan Kiros, and Geoffrey~E. Hinton.
\newblock Layer normalization.
\newblock \emph{CoRR}, abs/1607.06450, 2016.

\bibitem[Badia et~al.(2020)Badia, Piot, Kapturowski, Sprechmann, Vitvitskyi,
  Guo, and Blundell]{agent57}
Adri{\`{a}}~Puigdom{\`{e}}nech Badia, Bilal Piot, Steven Kapturowski, Pablo
  Sprechmann, Alex Vitvitskyi, Zhaohan~Daniel Guo, and Charles Blundell.
\newblock Agent57: Outperforming the atari human benchmark.
\newblock In \emph{Proceedings of the 37th International Conference on Machine
  Learning, {ICML} 2020, 13-18 July 2020, Virtual Event}, volume 119 of
  \emph{Proceedings of Machine Learning Research}, pp.\  507--517. {PMLR},
  2020.

\bibitem[Banijamali et~al.(2018)Banijamali, Shu, Bui, Ghodsi, et~al.]{rce}
Ershad Banijamali, Rui Shu, Hung Bui, Ali Ghodsi, et~al.
\newblock Robust locally-linear controllable embedding.
\newblock In \emph{International Conference on Artificial Intelligence and
  Statistics}, pp.\  1751--1759. PMLR, 2018.

\bibitem[Bardes et~al.(2022)Bardes, Ponce, and LeCun]{vicreg}
Adrien Bardes, Jean Ponce, and Yann LeCun.
\newblock Vicreg: Variance-invariance-covariance regularization for
  self-supervised learning.
\newblock In \emph{The Tenth International Conference on Learning
  Representations, {ICLR} 2022, Virtual Event, April 25-29, 2022}.
  OpenReview.net, 2022.

\bibitem[Bellemare et~al.(2013)Bellemare, Naddaf, Veness, and Bowling]{ale}
Marc~G. Bellemare, Yavar Naddaf, Joel Veness, and Michael Bowling.
\newblock The arcade learning environment: An evaluation platform for general
  agents.
\newblock \emph{J. Artif. Intell. Res.}, 47:\penalty0 253--279, 2013.
\newblock \doi{10.1613/jair.3912}.

\bibitem[Bengio et~al.(2013)Bengio, L{\'e}onard, and
  Courville]{straight-through}
Yoshua Bengio, Nicholas L{\'e}onard, and Aaron Courville.
\newblock Estimating or propagating gradients through stochastic neurons for
  conditional computation.
\newblock \emph{arXiv preprint arXiv:1308.3432}, 2013.

\bibitem[Bromley et~al.(1993)Bromley, Guyon, LeCun, S{\"{a}}ckinger, and
  Shah]{siamese}
Jane Bromley, Isabelle Guyon, Yann LeCun, Eduard S{\"{a}}ckinger, and Roopak
  Shah.
\newblock Signature verification using a siamese time delay neural network.
\newblock In \emph{Advances in Neural Information Processing Systems 6, [7th
  {NIPS} Conference, Denver, Colorado, USA, 1993]}, pp.\  737--744. Morgan
  Kaufmann, 1993.

\bibitem[Caron et~al.(2020)Caron, Misra, Mairal, Goyal, Bojanowski, and
  Joulin]{swav}
Mathilde Caron, Ishan Misra, Julien Mairal, Priya Goyal, Piotr Bojanowski, and
  Armand Joulin.
\newblock Unsupervised learning of visual features by contrasting cluster
  assignments.
\newblock In \emph{Advances in Neural Information Processing Systems 33: Annual
  Conference on Neural Information Processing Systems 2020, NeurIPS 2020,
  December 6-12, 2020, virtual}, 2020.

\bibitem[Caron et~al.(2021)Caron, Touvron, Misra, J{\'{e}}gou, Mairal,
  Bojanowski, and Joulin]{dino}
Mathilde Caron, Hugo Touvron, Ishan Misra, Herv{\'{e}} J{\'{e}}gou, Julien
  Mairal, Piotr Bojanowski, and Armand Joulin.
\newblock Emerging properties in self-supervised vision transformers.
\newblock In \emph{2021 {IEEE/CVF} International Conference on Computer Vision,
  {ICCV} 2021, Montreal, QC, Canada, October 10-17, 2021}, pp.\  9630--9640.
  {IEEE}, 2021.
\newblock \doi{10.1109/ICCV48922.2021.00951}.

\bibitem[Chen et~al.(2020)Chen, Kornblith, Norouzi, and Hinton]{simclr}
Ting Chen, Simon Kornblith, Mohammad Norouzi, and Geoffrey~E. Hinton.
\newblock A simple framework for contrastive learning of visual
  representations.
\newblock In \emph{Proceedings of the 37th International Conference on Machine
  Learning, {ICML} 2020, 13-18 July 2020, Virtual Event}, volume 119 of
  \emph{Proceedings of Machine Learning Research}, pp.\  1597--1607. {PMLR},
  2020.

\bibitem[Chen \& He(2021)Chen and He]{simsiam}
Xinlei Chen and Kaiming He.
\newblock Exploring simple siamese representation learning.
\newblock In \emph{{IEEE} Conference on Computer Vision and Pattern
  Recognition, {CVPR} 2021, virtual, June 19-25, 2021}, pp.\  15750--15758.
  Computer Vision Foundation / {IEEE}, 2021.
\newblock \doi{10.1109/CVPR46437.2021.01549}.

\bibitem[Chua et~al.(2018)Chua, Calandra, McAllister, and Levine]{pets}
Kurtland Chua, Roberto Calandra, Rowan McAllister, and Sergey Levine.
\newblock Deep reinforcement learning in a handful of trials using
  probabilistic dynamics models.
\newblock \emph{Advances in neural information processing systems}, 31, 2018.

\bibitem[Elias(1955)]{predictive-coding}
Peter Elias.
\newblock Predictive coding--i.
\newblock \emph{IRE transactions on information theory}, 1\penalty0
  (1):\penalty0 16--24, 1955.

\bibitem[Ferns \& Precup(2014)Ferns and Precup]{bisimulation-metrics}
Norman Ferns and Doina Precup.
\newblock Bisimulation metrics are optimal value functions.
\newblock In \emph{UAI}, pp.\  210--219, 2014.

\bibitem[Garrido et~al.(2022)Garrido, Chen, Bardes, Najman, and Lecun]{duality}
Quentin Garrido, Yubei Chen, Adrien Bardes, Laurent Najman, and Yann Lecun.
\newblock On the duality between contrastive and non-contrastive
  self-supervised learning.
\newblock \emph{arXiv preprint arXiv:2206.02574}, 2022.

\bibitem[Gelada et~al.(2019)Gelada, Kumar, Buckman, Nachum, and
  Bellemare]{deepmdp}
Carles Gelada, Saurabh Kumar, Jacob Buckman, Ofir Nachum, and Marc~G.
  Bellemare.
\newblock Deepmdp: Learning continuous latent space models for representation
  learning.
\newblock In \emph{Proceedings of the 36th International Conference on Machine
  Learning, {ICML} 2019, 9-15 June 2019, Long Beach, California, {USA}},
  volume~97 of \emph{Proceedings of Machine Learning Research}, pp.\
  2170--2179. {PMLR}, 2019.

\bibitem[Grill et~al.(2020)Grill, Strub, Altch{\'{e}}, Tallec, Richemond,
  Buchatskaya, Doersch, Pires, Guo, Azar, Piot, Kavukcuoglu, Munos, and
  Valko]{byol}
Jean{-}Bastien Grill, Florian Strub, Florent Altch{\'{e}}, Corentin Tallec,
  Pierre~H. Richemond, Elena Buchatskaya, Carl Doersch, Bernardo~{\'{A}}vila
  Pires, Zhaohan Guo, Mohammad~Gheshlaghi Azar, Bilal Piot, Koray Kavukcuoglu,
  R{\'{e}}mi Munos, and Michal Valko.
\newblock Bootstrap your own latent - {A} new approach to self-supervised
  learning.
\newblock In \emph{Advances in Neural Information Processing Systems 33: Annual
  Conference on Neural Information Processing Systems 2020, NeurIPS 2020,
  December 6-12, 2020, virtual}, 2020.

\bibitem[Ha \& Schmidhuber(2018)Ha and Schmidhuber]{world-models}
David Ha and J{\"{u}}rgen Schmidhuber.
\newblock Recurrent world models facilitate policy evolution.
\newblock In \emph{Advances in Neural Information Processing Systems 31: Annual
  Conference on Neural Information Processing Systems 2018, NeurIPS 2018,
  December 3-8, 2018, Montr{\'{e}}al, Canada}, pp.\  2455--2467, 2018.

\bibitem[Hafner et~al.(2019)Hafner, Lillicrap, Fischer, Villegas, Ha, Lee, and
  Davidson]{planet}
Danijar Hafner, Timothy~P. Lillicrap, Ian Fischer, Ruben Villegas, David Ha,
  Honglak Lee, and James Davidson.
\newblock Learning latent dynamics for planning from pixels.
\newblock In \emph{Proceedings of the 36th International Conference on Machine
  Learning, {ICML} 2019, 9-15 June 2019, Long Beach, California, {USA}},
  volume~97 of \emph{Proceedings of Machine Learning Research}, pp.\
  2555--2565. {PMLR}, 2019.

\bibitem[Hafner et~al.(2020)Hafner, Lillicrap, Ba, and Norouzi]{dreamerv1}
Danijar Hafner, Timothy~P. Lillicrap, Jimmy Ba, and Mohammad Norouzi.
\newblock Dream to control: Learning behaviors by latent imagination.
\newblock In \emph{8th International Conference on Learning Representations,
  {ICLR} 2020, Addis Ababa, Ethiopia, April 26-30, 2020}. OpenReview.net, 2020.

\bibitem[Hafner et~al.(2021)Hafner, Lillicrap, Norouzi, and Ba]{dreamerv2}
Danijar Hafner, Timothy~P. Lillicrap, Mohammad Norouzi, and Jimmy Ba.
\newblock Mastering atari with discrete world models.
\newblock In \emph{9th International Conference on Learning Representations,
  {ICLR} 2021, Virtual Event, Austria, May 3-7, 2021}. OpenReview.net, 2021.

\bibitem[Hafner et~al.(2023)Hafner, Pasukonis, Ba, and Lillicrap]{dreamerv3}
Danijar Hafner, Jurgis Pasukonis, Jimmy Ba, and Timothy~P. Lillicrap.
\newblock Mastering diverse domains through world models.
\newblock \emph{CoRR}, abs/2301.04104, 2023.
\newblock \doi{10.48550/arXiv.2301.04104}.

\bibitem[Hansen et~al.(2022)Hansen, Su, and Wang]{td-mpc}
Nicklas Hansen, Hao Su, and Xiaolong Wang.
\newblock Temporal difference learning for model predictive control.
\newblock In \emph{International Conference on Machine Learning, {ICML} 2022,
  17-23 July 2022, Baltimore, Maryland, {USA}}, volume 162 of \emph{Proceedings
  of Machine Learning Research}, pp.\  8387--8406. {PMLR}, 2022.

\bibitem[He et~al.(2016)He, Zhang, Ren, and Sun]{resnet}
Kaiming He, Xiangyu Zhang, Shaoqing Ren, and Jian Sun.
\newblock Deep residual learning for image recognition.
\newblock In \emph{Proceedings of the IEEE conference on computer vision and
  pattern recognition}, pp.\  770--778, 2016.

\bibitem[He et~al.(2020)He, Fan, Wu, Xie, and Girshick]{moco}
Kaiming He, Haoqi Fan, Yuxin Wu, Saining Xie, and Ross~B. Girshick.
\newblock Momentum contrast for unsupervised visual representation learning.
\newblock In \emph{2020 {IEEE/CVF} Conference on Computer Vision and Pattern
  Recognition, {CVPR} 2020, Seattle, WA, USA, June 13-19, 2020}, pp.\
  9726--9735. Computer Vision Foundation / {IEEE}, 2020.
\newblock \doi{10.1109/CVPR42600.2020.00975}.

\bibitem[H{\'e}naff et~al.(2019)H{\'e}naff, Goris, and
  Simoncelli]{perceptual-straightening}
Olivier~J H{\'e}naff, Robbe~LT Goris, and Eero~P Simoncelli.
\newblock Perceptual straightening of natural videos.
\newblock \emph{Nature neuroscience}, 22\penalty0 (6):\penalty0 984--991, 2019.

\bibitem[H{\'{e}}naff et~al.(2019)H{\'{e}}naff, Srinivas, Fauw, Razavi,
  Doersch, Eslami, and van~den Oord]{cpcv2}
Olivier~J. H{\'{e}}naff, Aravind Srinivas, Jeffrey~De Fauw, Ali Razavi, Carl
  Doersch, S.~M.~Ali Eslami, and A{\"{a}}ron van~den Oord.
\newblock Data-efficient image recognition with contrastive predictive coding.
\newblock \emph{CoRR}, abs/1905.09272, 2019.
\newblock URL \url{http://arxiv.org/abs/1905.09272}.

\bibitem[Hendrycks \& Gimpel(2016)Hendrycks and Gimpel]{gelu}
Dan Hendrycks and Kevin Gimpel.
\newblock Bridging nonlinearities and stochastic regularizers with gaussian
  error linear units.
\newblock \emph{CoRR}, abs/1606.08415, 2016.

\bibitem[Hessel et~al.(2018)Hessel, Modayil, van Hasselt, Schaul, Ostrovski,
  Dabney, Horgan, Piot, Azar, and Silver]{rainbow}
Matteo Hessel, Joseph Modayil, Hado van Hasselt, Tom Schaul, Georg Ostrovski,
  Will Dabney, Dan Horgan, Bilal Piot, Mohammad~Gheshlaghi Azar, and David
  Silver.
\newblock Rainbow: Combining improvements in deep reinforcement learning.
\newblock In \emph{Proceedings of the Thirty-Second {AAAI} Conference on
  Artificial Intelligence, (AAAI-18), the 30th innovative Applications of
  Artificial Intelligence (IAAI-18), and the 8th {AAAI} Symposium on
  Educational Advances in Artificial Intelligence (EAAI-18), New Orleans,
  Louisiana, USA, February 2-7, 2018}, pp.\  3215--3222. {AAAI} Press, 2018.

\bibitem[Hosoya et~al.(2005)Hosoya, Baccus, and Meister]{predictive-retina}
Toshihiko Hosoya, Stephen~A Baccus, and Markus Meister.
\newblock Dynamic predictive coding by the retina.
\newblock \emph{Nature}, 436\penalty0 (7047):\penalty0 71--77, 2005.

\bibitem[Huang \& Rao(2011)Huang and Rao]{predictive-review}
Yanping Huang and Rajesh~PN Rao.
\newblock Predictive coding.
\newblock \emph{Wiley Interdisciplinary Reviews: Cognitive Science}, 2\penalty0
  (5):\penalty0 580--593, 2011.

\bibitem[Ioffe \& Szegedy(2015)Ioffe and Szegedy]{batch-norm}
Sergey Ioffe and Christian Szegedy.
\newblock Batch normalization: Accelerating deep network training by reducing
  internal covariate shift.
\newblock In \emph{Proceedings of the 32nd International Conference on Machine
  Learning, {ICML} 2015, Lille, France, 6-11 July 2015}, volume~37 of
  \emph{{JMLR} Workshop and Conference Proceedings}, pp.\  448--456. JMLR.org,
  2015.

\bibitem[Kaiser et~al.(2020)Kaiser, Babaeizadeh, Milos, Osinski, Campbell,
  Czechowski, Erhan, Finn, Kozakowski, Levine, Mohiuddin, Sepassi, Tucker, and
  Michalewski]{simple}
Lukasz Kaiser, Mohammad Babaeizadeh, Piotr Milos, Blazej Osinski, Roy~H.
  Campbell, Konrad Czechowski, Dumitru Erhan, Chelsea Finn, Piotr Kozakowski,
  Sergey Levine, Afroz Mohiuddin, Ryan Sepassi, George Tucker, and Henryk
  Michalewski.
\newblock Model based reinforcement learning for atari.
\newblock In \emph{8th International Conference on Learning Representations,
  {ICLR} 2020, Addis Ababa, Ethiopia, April 26-30, 2020}. OpenReview.net, 2020.

\bibitem[Kapturowski et~al.(2023)Kapturowski, Campos, Jiang, Rakicevic, van
  Hasselt, Blundell, and Badia]{meme}
Steven Kapturowski, Victor Campos, Ray Jiang, Nemanja Rakicevic, Hado van
  Hasselt, Charles Blundell, and Adri{\`{a}}~Puigdom{\`{e}}nech Badia.
\newblock Human-level atari 200x faster.
\newblock In \emph{The Eleventh International Conference on Learning
  Representations, {ICLR} 2023, Kigali, Rwanda, May 1-5, 2023}. OpenReview.net,
  2023.

\bibitem[Laskin et~al.(2020{\natexlab{a}})Laskin, Lee, Stooke, Pinto, Abbeel,
  and Srinivas]{rad}
Michael Laskin, Kimin Lee, Adam Stooke, Lerrel Pinto, Pieter Abbeel, and
  Aravind Srinivas.
\newblock Reinforcement learning with augmented data.
\newblock In \emph{Advances in Neural Information Processing Systems 33: Annual
  Conference on Neural Information Processing Systems 2020, NeurIPS 2020,
  December 6-12, 2020, virtual}, 2020{\natexlab{a}}.

\bibitem[Laskin et~al.(2020{\natexlab{b}})Laskin, Srinivas, and Abbeel]{curl}
Michael Laskin, Aravind Srinivas, and Pieter Abbeel.
\newblock {CURL:} contrastive unsupervised representations for reinforcement
  learning.
\newblock In \emph{Proceedings of the 37th International Conference on Machine
  Learning, {ICML} 2020, 13-18 July 2020, Virtual Event}, volume 119 of
  \emph{Proceedings of Machine Learning Research}, pp.\  5639--5650. {PMLR},
  2020{\natexlab{b}}.

\bibitem[LeCun(2022)]{path}
Yann LeCun.
\newblock A path towards autonomous machine intelligence version 0.9. 2,
  2022-06-27.
\newblock \emph{Open Review}, 62\penalty0 (1), 2022.

\bibitem[Lee et~al.(2020)Lee, Nagabandi, Abbeel, and Levine]{slac}
Alex~X Lee, Anusha Nagabandi, Pieter Abbeel, and Sergey Levine.
\newblock Stochastic latent actor-critic: Deep reinforcement learning with a
  latent variable model.
\newblock \emph{Advances in Neural Information Processing Systems},
  33:\penalty0 741--752, 2020.

\bibitem[Loshchilov \& Hutter(2019)Loshchilov and Hutter]{adamw}
Ilya Loshchilov and Frank Hutter.
\newblock Decoupled weight decay regularization.
\newblock In \emph{7th International Conference on Learning Representations,
  {ICLR} 2019, New Orleans, LA, USA, May 6-9, 2019}. OpenReview.net, 2019.

\bibitem[Machado et~al.(2018)Machado, Bellemare, Talvitie, Veness, Hausknecht,
  and Bowling]{revisiting}
Marlos~C Machado, Marc~G Bellemare, Erik Talvitie, Joel Veness, Matthew
  Hausknecht, and Michael Bowling.
\newblock Revisiting the arcade learning environment: Evaluation protocols and
  open problems for general agents.
\newblock \emph{Journal of Artificial Intelligence Research}, 61:\penalty0
  523--562, 2018.

\bibitem[Micheli et~al.(2023)Micheli, Alonso, and Fleuret]{iris}
Vincent Micheli, Eloi Alonso, and Fran{\c{c}}ois Fleuret.
\newblock Transformers are sample-efficient world models.
\newblock In \emph{The Eleventh International Conference on Learning
  Representations, {ICLR} 2023, Kigali, Rwanda, May 1-5, 2023}. OpenReview.net,
  2023.

\bibitem[Mnih et~al.(2015)Mnih, Kavukcuoglu, Silver, Rusu, Veness, Bellemare,
  Graves, Riedmiller, Fidjeland, Ostrovski, Petersen, Beattie, Sadik,
  Antonoglou, King, Kumaran, Wierstra, Legg, and Hassabis]{dqn}
Volodymyr Mnih, Koray Kavukcuoglu, David Silver, Andrei~A. Rusu, Joel Veness,
  Marc~G. Bellemare, Alex Graves, Martin~A. Riedmiller, Andreas Fidjeland,
  Georg Ostrovski, Stig Petersen, Charles Beattie, Amir Sadik, Ioannis
  Antonoglou, Helen King, Dharshan Kumaran, Daan Wierstra, Shane Legg, and
  Demis Hassabis.
\newblock Human-level control through deep reinforcement learning.
\newblock \emph{Nat.}, 518\penalty0 (7540):\penalty0 529--533, 2015.
\newblock \doi{10.1038/nature14236}.

\bibitem[Mnih et~al.(2016)Mnih, Badia, Mirza, Graves, Lillicrap, Harley,
  Silver, and Kavukcuoglu]{a3c}
Volodymyr Mnih, Adri{\`{a}}~Puigdom{\`{e}}nech Badia, Mehdi Mirza, Alex Graves,
  Timothy~P. Lillicrap, Tim Harley, David Silver, and Koray Kavukcuoglu.
\newblock Asynchronous methods for deep reinforcement learning.
\newblock In \emph{Proceedings of the 33nd International Conference on Machine
  Learning, {ICML} 2016, New York City, NY, USA, June 19-24, 2016}, volume~48
  of \emph{{JMLR} Workshop and Conference Proceedings}, pp.\  1928--1937.
  JMLR.org, 2016.

\bibitem[Oh et~al.(2017)Oh, Singh, and Lee]{vp-net}
Junhyuk Oh, Satinder Singh, and Honglak Lee.
\newblock Value prediction network.
\newblock \emph{Advances in neural information processing systems}, 30, 2017.

\bibitem[Oord et~al.(2018)Oord, Li, and Vinyals]{cpc}
Aaron van~den Oord, Yazhe Li, and Oriol Vinyals.
\newblock Representation learning with contrastive predictive coding.
\newblock \emph{arXiv preprint arXiv:1807.03748}, 2018.

\bibitem[Rao \& Ballard(1999)Rao and Ballard]{predictive-visual}
Rajesh~PN Rao and Dana~H Ballard.
\newblock Predictive coding in the visual cortex: a functional interpretation
  of some extra-classical receptive-field effects.
\newblock \emph{Nature neuroscience}, 2\penalty0 (1):\penalty0 79--87, 1999.

\bibitem[Robine et~al.(2023)Robine, H{\"{o}}ftmann, Uelwer, and Harmeling]{twm}
Jan Robine, Marc H{\"{o}}ftmann, Tobias Uelwer, and Stefan Harmeling.
\newblock Transformer-based world models are happy with 100k interactions.
\newblock In \emph{The Eleventh International Conference on Learning
  Representations, {ICLR} 2023, Kigali, Rwanda, May 1-5, 2023}. OpenReview.net,
  2023.

\bibitem[Schrittwieser et~al.(2020)Schrittwieser, Antonoglou, Hubert, Simonyan,
  Sifre, Schmitt, Guez, Lockhart, Hassabis, Graepel, Lillicrap, and
  Silver]{muzero}
Julian Schrittwieser, Ioannis Antonoglou, Thomas Hubert, Karen Simonyan,
  Laurent Sifre, Simon Schmitt, Arthur Guez, Edward Lockhart, Demis Hassabis,
  Thore Graepel, Timothy~P. Lillicrap, and David Silver.
\newblock Mastering atari, go, chess and shogi by planning with a learned
  model.
\newblock \emph{Nat.}, 588\penalty0 (7839):\penalty0 604--609, 2020.
\newblock \doi{10.1038/s41586-020-03051-4}.

\bibitem[Schulman et~al.(2016)Schulman, Moritz, Levine, Jordan, and
  Abbeel]{gae}
John Schulman, Philipp Moritz, Sergey Levine, Michael~I. Jordan, and Pieter
  Abbeel.
\newblock High-dimensional continuous control using generalized advantage
  estimation.
\newblock In \emph{4th International Conference on Learning Representations,
  {ICLR} 2016, San Juan, Puerto Rico, May 2-4, 2016, Conference Track
  Proceedings}, 2016.

\bibitem[Schulman et~al.(2017)Schulman, Wolski, Dhariwal, Radford, and
  Klimov]{ppo}
John Schulman, Filip Wolski, Prafulla Dhariwal, Alec Radford, and Oleg Klimov.
\newblock Proximal policy optimization algorithms.
\newblock \emph{CoRR}, abs/1707.06347, 2017.

\bibitem[Schwarzer et~al.(2021{\natexlab{a}})Schwarzer, Anand, Goel, Hjelm,
  Courville, and Bachman]{spr}
Max Schwarzer, Ankesh Anand, Rishab Goel, R.~Devon Hjelm, Aaron~C. Courville,
  and Philip Bachman.
\newblock Data-efficient reinforcement learning with self-predictive
  representations.
\newblock In \emph{9th International Conference on Learning Representations,
  {ICLR} 2021, Virtual Event, Austria, May 3-7, 2021}. OpenReview.net,
  2021{\natexlab{a}}.

\bibitem[Schwarzer et~al.(2021{\natexlab{b}})Schwarzer, Rajkumar, Noukhovitch,
  Anand, Charlin, Hjelm, Bachman, and Courville]{spi}
Max Schwarzer, Nitarshan Rajkumar, Michael Noukhovitch, Ankesh Anand, Laurent
  Charlin, R.~Devon Hjelm, Philip Bachman, and Aaron~C. Courville.
\newblock Pretraining representations for data-efficient reinforcement
  learning.
\newblock \emph{CoRR}, abs/2106.04799, 2021{\natexlab{b}}.

\bibitem[Schwarzer et~al.(2023)Schwarzer, Obando{-}Ceron, Courville, Bellemare,
  Agarwal, and Castro]{bbf}
Max Schwarzer, Johan~Samir Obando{-}Ceron, Aaron~C. Courville, Marc~G.
  Bellemare, Rishabh Agarwal, and Pablo~Samuel Castro.
\newblock Bigger, better, faster: Human-level atari with human-level
  efficiency.
\newblock In \emph{International Conference on Machine Learning, {ICML} 2023,
  23-29 July 2023, Honolulu, Hawaii, {USA}}, volume 202 of \emph{Proceedings of
  Machine Learning Research}, pp.\  30365--30380. {PMLR}, 2023.

\bibitem[Silver et~al.(2017)Silver, Hasselt, Hessel, Schaul, Guez, Harley,
  Dulac-Arnold, Reichert, Rabinowitz, Barreto, et~al.]{predictron}
David Silver, Hado Hasselt, Matteo Hessel, Tom Schaul, Arthur Guez, Tim Harley,
  Gabriel Dulac-Arnold, David Reichert, Neil Rabinowitz, Andre Barreto, et~al.
\newblock The predictron: End-to-end learning and planning.
\newblock In \emph{International Conference on Machine Learning}, pp.\
  3191--3199. PMLR, 2017.

\bibitem[Sutton(1991)]{dyna}
Richard~S. Sutton.
\newblock Dyna, an integrated architecture for learning, planning, and
  reacting.
\newblock \emph{{SIGART} Bull.}, 2\penalty0 (4):\penalty0 160--163, 1991.
\newblock \doi{10.1145/122344.122377}.

\bibitem[Sutton \& Barto(2018)Sutton and Barto]{rl-intro}
Richard~S. Sutton and Andrew~G. Barto.
\newblock \emph{Reinforcement Learning: An Introduction}.
\newblock A Bradford Book, Cambridge, MA, USA, 2018.
\newblock ISBN 0262039249.

\bibitem[Sutton et~al.(1999)Sutton, McAllester, Singh, and
  Mansour]{policy-gradient}
Richard~S. Sutton, David~A. McAllester, Satinder Singh, and Yishay Mansour.
\newblock Policy gradient methods for reinforcement learning with function
  approximation.
\newblock In \emph{Advances in Neural Information Processing Systems 12,
  {[NIPS} Conference, Denver, Colorado, USA, November 29 - December 4, 1999]},
  pp.\  1057--1063. The {MIT} Press, 1999.

\bibitem[Tamar et~al.(2016)Tamar, Wu, Thomas, Levine, and Abbeel]{vi-net}
Aviv Tamar, Yi~Wu, Garrett Thomas, Sergey Levine, and Pieter Abbeel.
\newblock Value iteration networks.
\newblock \emph{Advances in neural information processing systems}, 29, 2016.

\bibitem[Van~der Maaten \& Hinton(2008)Van~der Maaten and Hinton]{tsne}
Laurens Van~der Maaten and Geoffrey Hinton.
\newblock Visualizing data using t-sne.
\newblock \emph{Journal of machine learning research}, 9\penalty0 (11), 2008.

\bibitem[van Hasselt et~al.(2019)van Hasselt, Hessel, and Aslanides]{der}
Hado van Hasselt, Matteo Hessel, and John Aslanides.
\newblock When to use parametric models in reinforcement learning?
\newblock In \emph{Advances in Neural Information Processing Systems 32: Annual
  Conference on Neural Information Processing Systems 2019, NeurIPS 2019,
  December 8-14, 2019, Vancouver, BC, Canada}, pp.\  14322--14333, 2019.

\bibitem[Watter et~al.(2015)Watter, Springenberg, Boedecker, and
  Riedmiller]{e2c}
Manuel Watter, Jost Springenberg, Joschka Boedecker, and Martin Riedmiller.
\newblock Embed to control: A locally linear latent dynamics model for control
  from raw images.
\newblock \emph{Advances in neural information processing systems}, 28, 2015.

\bibitem[Williams \& Peng(1991)Williams and Peng]{rl-entropy}
Ronald~J. Williams and Jing Peng.
\newblock Function optimization using connectionist reinforcement learning
  algorithms.
\newblock \emph{Connection Science}, 3\penalty0 (3):\penalty0 241--268, 1991.
\newblock \doi{10.1080/09540099108946587}.

\bibitem[Yarats et~al.(2021)Yarats, Kostrikov, and Fergus]{drq}
Denis Yarats, Ilya Kostrikov, and Rob Fergus.
\newblock Image augmentation is all you need: Regularizing deep reinforcement
  learning from pixels.
\newblock In \emph{9th International Conference on Learning Representations,
  {ICLR} 2021, Virtual Event, Austria, May 3-7, 2021}. OpenReview.net, 2021.

\bibitem[Ye et~al.(2021)Ye, Liu, Kurutach, Abbeel, and Gao]{efficient-zero}
Weirui Ye, Shaohuai Liu, Thanard Kurutach, Pieter Abbeel, and Yang Gao.
\newblock Mastering atari games with limited data.
\newblock In \emph{Advances in Neural Information Processing Systems 34: Annual
  Conference on Neural Information Processing Systems 2021, NeurIPS 2021,
  December 6-14, 2021, virtual}, pp.\  25476--25488, 2021.

\bibitem[Zbontar et~al.(2021)Zbontar, Jing, Misra, LeCun, and Deny]{barlow}
Jure Zbontar, Li~Jing, Ishan Misra, Yann LeCun, and St{\'{e}}phane Deny.
\newblock Barlow twins: Self-supervised learning via redundancy reduction.
\newblock In \emph{Proceedings of the 38th International Conference on Machine
  Learning, {ICML} 2021, 18-24 July 2021, Virtual Event}, volume 139 of
  \emph{Proceedings of Machine Learning Research}, pp.\  12310--12320. {PMLR},
  2021.

\bibitem[Zhang et~al.(2021)Zhang, McAllister, Calandra, Gal, and Levine]{dbc}
Amy Zhang, Rowan~Thomas McAllister, Roberto Calandra, Yarin Gal, and Sergey
  Levine.
\newblock Learning invariant representations for reinforcement learning without
  reconstruction.
\newblock In \emph{9th International Conference on Learning Representations,
  {ICLR} 2021, Virtual Event, Austria, May 3-7, 2021}. OpenReview.net, 2021.

\end{thebibliography}
\bibliographystyle{iclr2025_conference}

\newpage
\appendix

\section{Ethics Statement}
\label{sec:impact}

The advancements in developing Simple, Good, and Fast (SGF) world models for
reinforcement learning can significantly enhance various fields by making
advanced techniques more accessible and reducing computational demands. While
this democratization can drive innovation in areas like robotics and autonomous
systems, it also raises ethical concerns, such as potential misuse in
surveillance or autonomous weaponry. Therefore, it is crucial for the research
community to address these risks and develop guidelines to ensure responsible
use.


\section{Relations to Other Methods} \label{sec:add-relations}

\subsection{Reconstruction-Free Models} \label{sec:rec-free}

Previous world models learning in imagination rely on image reconstructions.
However, there are other approaches that learn a model without relying on image
reconstructions, which we will discuss in this section.


\paragraph{Value equivalence.}
In the \emph{value equivalence} paradigm, trajectories of the model must
achieve the same cumulative rewards as those in the real environment,
regardless of whether the produced hidden states correspond to any real
environment states or not. These models are used for decision-time planning.
For instance, MuZero \citep{muzero} trains a model with hidden states by
predicting the policy, the value function, and the reward. EfficientZero
\citep{efficient-zero} extends this objective by introducing a self-supervised
consistency loss with a projector and a predictor network, which shares
similarities with our temporal consistency loss. However, they predict the next
representation before feeding it into the projector, rather than employing an
action-conditioned predictor. Additionally, their approach bears more
resemblance to SimSiam \citep{simsiam}, as they utilize the stop-gradient
operation and lack explicit information maximization.

\paragraph{Auxiliary tasks and bisimulation metrics.}
Learning a model as an auxiliary task can improve the representations of the
model-free agent. Although these approaches only share a loose connection to
generative world models, they typically are reconstruction-free. For instance,
SPR \citep{spr} learns a transition model to predict the latent states of future
time steps using a projector and a predictor network, incorporating data
augmentation. Their architecture shares similarities with ours, however, their
transition model is convolutional, and they predict the next representation
before feeding it into the projector (similar to EfficientZero). Moreover, their
methodology is influenced by BYOL \citep{byol}, utilizing a momentum encoder and
lacking explicit information maximization. SPI \citep{spi} combines SPR with
goal-conditioned RL and inverse dynamics modelling, i.e., predicting the action
$a_t$ from states $s_t$ and $s_{t+1}$. They pretrain an encoder on unlabelled
data, which is later finetuned on task-specific data.

A special type of auxiliary tasks are connected to bisimulation metrics, where
``behaviorally similar'' states are grouped together
\citep{bisimulation-metrics}. Similar to our approach, this also amounts to
learning a reward model and a (distributional) latent transition model (by
minimizing the Wasserstein distance). Prominent works include DeepMDP
\citep{deepmdp}, which still requires reconstructions for good results on
Atari, and DBC \citep{dbc}.


\subsection{Self-Supervised Representation Learning} \label{sec:ssl}

Our self-supervised representation learning framework is similar to existing
visual representation learning methods. We describe the differences to the most
related works. Note that a common difference is our architecture, as we employ
layer normalization instead of batch normalization \citep{batch-norm}, SiLU
instead of ReLU nonlinearities, and no ResNet-based encoder \citep{resnet}.

\paragraph{Relation to VICReg {\normalfont \citep{vicreg}}.} Our work is greatly
inspired by VICReg, which is used to learn representations of (stationary)
images. Originally, the same image is augmented and fed into both branches of
the Siamese neural network. We augment two successive image observations, so the
two branches get different inputs, which are nonetheless related. Furthermore,
VICReg maximizes the similarity between the embeddings ${\vz}$ and ${\vz'}$
directly, whereas we integrate an action-conditioned predictor network, since
the observations ${\vo}$ and ${\vo'}$ might have a more complicated connection.

\paragraph{Relation to BYOL {\normalfont \citep{byol}} and SimSiam {\normalfont
      \citep{simsiam}}.}
The basis of our self-supervised learning setup is a combination of ideas from
VICReg and BYOL. Specifically, we adopt the predictor network concept from BYOL
but utilize the regularization terms from VICReg, omitting BYOL's momentum
encoder. Additionally, our predictor network is action-conditioned. Similarly,
our method is related to SimSiam \citep{simsiam}, which also employs a predictor
network, but we do not need the stop-gradient operation.

From a practical perspective, these methods could likely achieve comparable
performance with appropriate hyperparameter tuning. However, our decision to use
VICReg was motivated by its conceptual advantages, which we believe make it
particularly suitable for our framework: Specifically, VICReg offers two key
advantages over BYOL and SimSiam. First, it avoids the need for additional
target networks updated through moving averages (as in BYOL). Second, VICReg has
a more established theoretical foundation for its loss functions, leveraging
variance and covariance regularization to prevent representation collapse. In
contrast, BYOL and SimSiam rely on mechanisms like target networks or
stop-gradient operations, which are more heuristic in nature.

\subsection{Model-free Data Augmentation} \label{sec:model-free}

Our method is model-based, but there are model-free methods that also use data
augmentation. \citet{rad} also use data augmentation to increase the sample
efficiency, but by augmenting the image observations passed to the model-free
agent. \citet{drq} additionally regularize the value function such that it is
invariant to the augmentations.



\clearpage

\section{Additional Analysis} \label{sec:add-analysis}

\begin{figure}[H]
  \centering
  \begin{subfigure}[b]{0.45\textwidth}
    \centering
    \includegraphics[width=\textwidth,trim={2.25cm 1.5cm 2cm 1.75cm},clip]{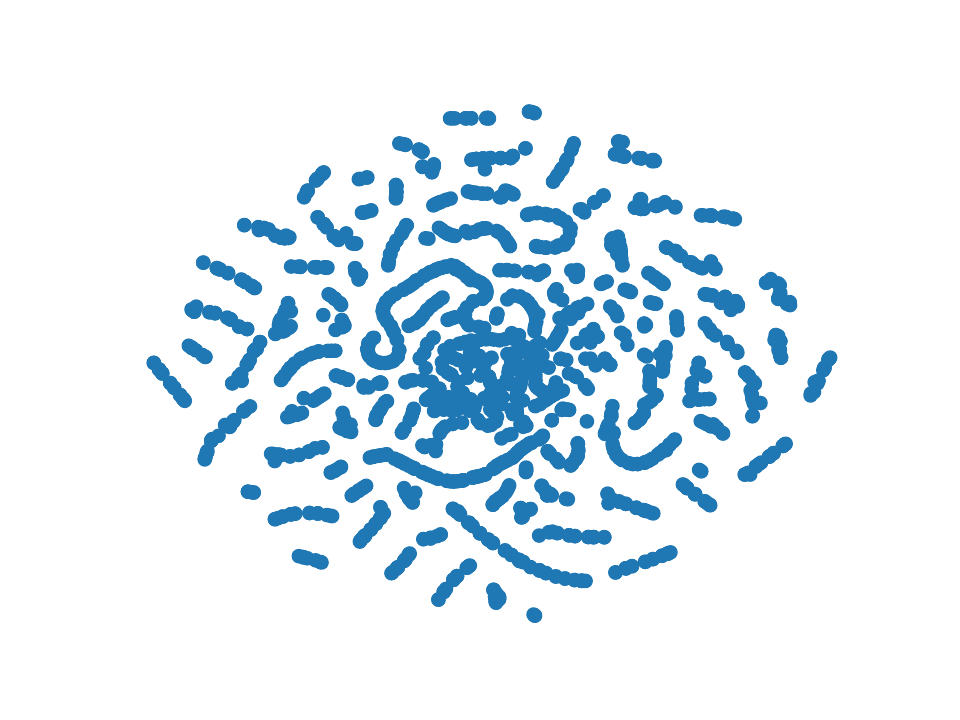}
    \caption{With temporal consistency.}
  \end{subfigure}
  \hspace{1cm}
  \begin{subfigure}[b]{0.45\textwidth}
    \centering
    \includegraphics[width=\textwidth,trim={2.25cm 1.5cm 2cm 1.75cm},clip]{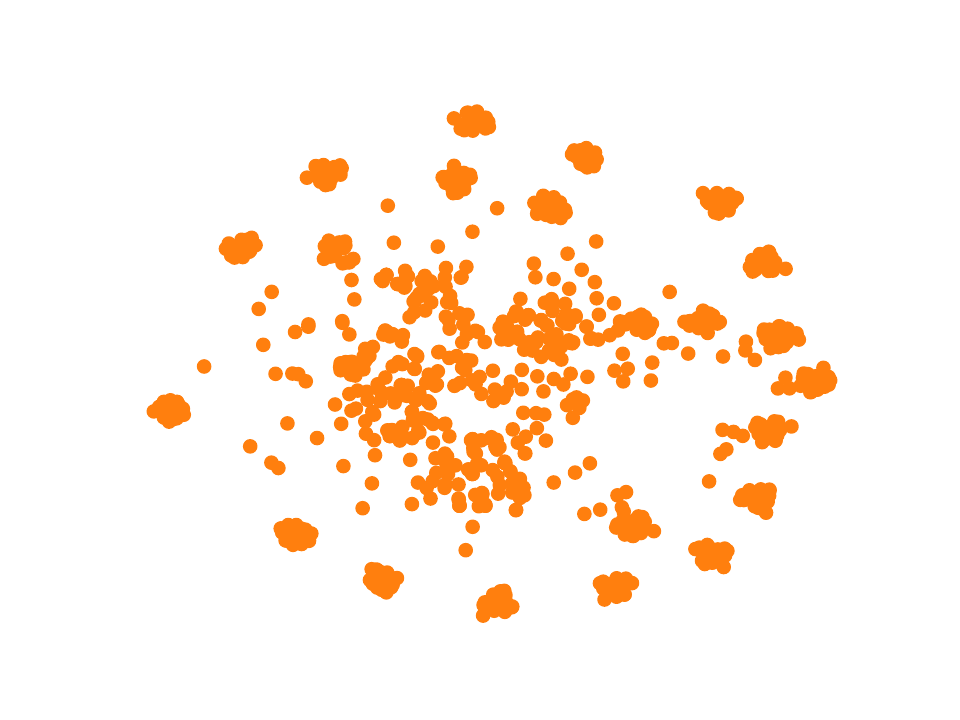}
    \caption{Without temporal consistency.}
  \end{subfigure}
  \caption{Comparison of two-dimensional t-SNE embeddings of the learned representations with and without
    temporal consistency. When only maximizing information, the representations are arranged in Gaussian
    blobs, which are harder to predict. We show the representations of one episode in Pong.}
  \label{fig:analyze-consistency}
\end{figure}

\begin{figure}[H]
  \centering
  \begin{subfigure}{\linewidth}
    \includegraphics[width=\linewidth]{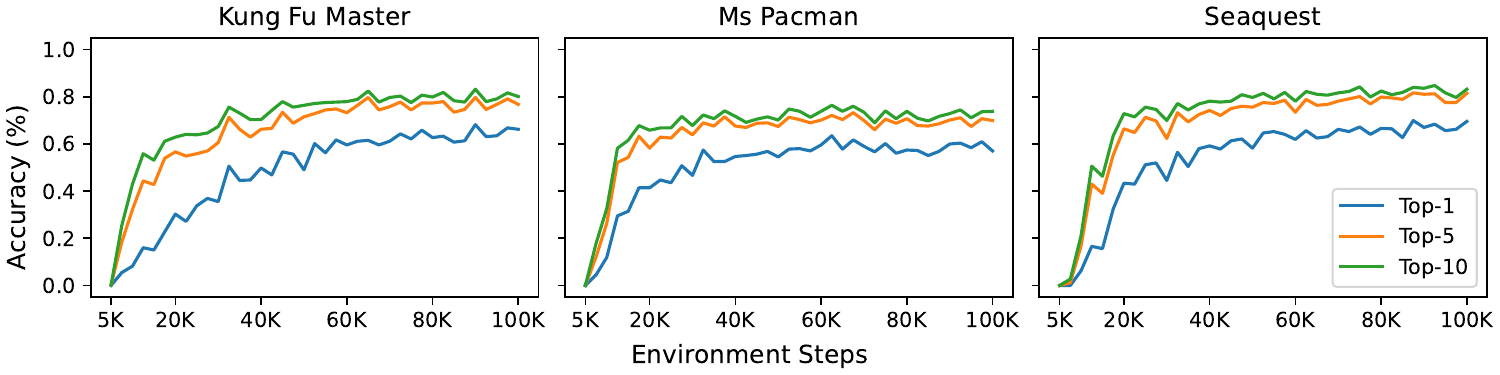}
    \caption{Top-$k$ accuracies of the reconstructions for different values of
      $k$. The accuracy is determined by first encoding and reconstructing an
      observation, then computing the MSE between the reconstruction and all
      ground truth observations in the replay buffer, and finally testing
      whether the input observation is among the $k$-nearest neighbors. We
      calculate the mean over a batch of $512$ observations. Note that the
      observations in the replay buffer can be very similar or even identical,
      so the top-$1$ accuracy is not as expressive as top-$5$ and top-$10$.}
  \end{subfigure}
  \vskip 3mm
  \begin{subfigure}{\linewidth}
    \includegraphics[width=\linewidth]{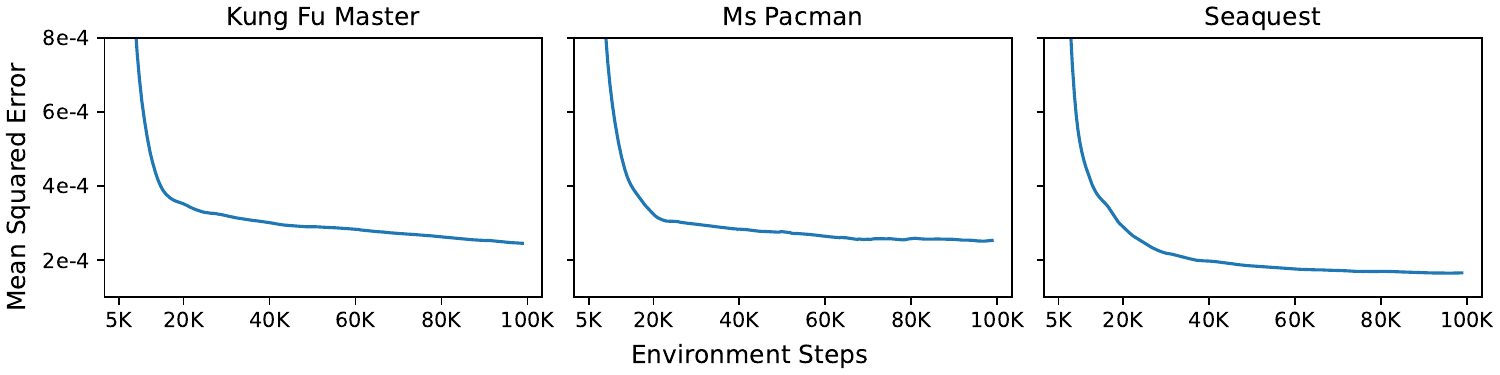}
    \caption{Reconstruction loss of the decoder.}
  \end{subfigure}
  \caption{Additional analysis of the decoder from \cref{sec:analysis}.
    We start training after collecting 5000 environment steps.}
\end{figure}

\clearpage

\section{Detailed Results}

\begin{table}[H]
  \caption{Comparison with other methods on the Atari100k benchmark. Averaged over 10 seeds.}
  \label{tab:results}
  \vspace{\belowcaptionskip}
  \small
  \centerline{
    \begin{tabularx}{1.1\linewidth}{l@{}ZZZZZZZl}
      \toprule
                        &                        &                & \multicolumn{1}{c}{\scriptsize Model-free} & \multicolumn{1}{c}{\scriptsize Lookahead} & \multicolumn{4}{c}{\scriptsize Learning in imagination}                                                                                 \\
      \cmidrule(lr){4-4} \cmidrule(lr){5-5} \cmidrule(lr){6-9}
      \textbf{Game}     & \mbox{\textbf{Random}} & \textbf{Human} & \multicolumn{1}{c}{\textbf{SPR}}           & \multicolumn{1}{c}{\textbf{Eff. Zero}}    & \multicolumn{1}{c}{\textbf{IRIS}}                       & \multicolumn{1}{c}{\textbf{DreamerV3}} & \textbf{\methodname{}} & (std. dev.) \\
      \midrule
      Alien             & 227.8                  & 7127.7         & 841.9                                      & 808.5                                     & 420.0                                                   & \textbf{959}                           & 518.8                  & (116.7)     \\
      Amidar            & 5.8                    & 1719.5         & \textbf{179.7}                             & 148.6                                     & 143.0                                                   & 139                                    & 62.7                   & (19.4)      \\
      Assault           & 222.4                  & 742.0          & 565.6                                      & 1263.1                                    & \textbf{1524.4}                                         & 706                                    & 850.1                  & (250.3)     \\
      Asterix           & 210.0                  & 8503.3         & 962.5                                      & \textbf{25557.8}                          & 853.6                                                   & 932                                    & 802.5                  & (317.5)     \\
      Bank Heist        & 14.2                   & 753.1          & 345.4                                      & 351.0                                     & 53.1                                                    & \textbf{649}                           & 58.7                   & (38.9)      \\
      Battle Zone       & 2360.0                 & 37187.5        & \textbf{14834.1}                           & 13871.2                                   & 13074.0                                                 & 12250                                  & 3747.0                 & (1240.3)    \\
      Boxing            & 0.1                    & 12.1           & 35.7                                       & 52.7                                      & 70.1                                                    & 78                                     & \textbf{83.4}          & (10.7)      \\
      Breakout          & 1.7                    & 30.5           & 19.6                                       & \textbf{414.1}                            & 83.7                                                    & 31                                     & 50.7                   & (37.6)      \\
      Chopper Cmd.      & 811.0                  & 7387.8         & 946.3                                      & 1117.3                                    & 1565.0                                                  & 420                                    & \textbf{1775.4}        & (593.9)     \\
      Crazy Climber     & 10780.5                & 35829.4        & 36700.5                                    & 83940.2                                   & 59324.2                                                 & \textbf{97190}                         & 15751.3                & (5488.9)    \\
      Demon Attack      & 152.1                  & 1971.0         & 517.6                                      & \textbf{13003.9}                          & 2034.4                                                  & 303                                    & 2809.5                 & (749.2)     \\
      Freeway           & 0.0                    & 29.6           & 19.3                                       & 21.8                                      & \textbf{31.1}                                           & 0                                      & 11.9                   & (4.7)       \\
      Frostbite         & 65.2                   & 4334.7         & \textbf{1170.7}                            & 296.3                                     & 259.1                                                   & 909                                    & 265.6                  & (8.6)       \\
      Gopher            & 257.6                  & 2412.5         & 660.6                                      & 3260.3                                    & 2236.1                                                  & \textbf{3730}                          & 416.4                  & (133.7)     \\
      Hero              & 1027.0                 & 30826.4        & 5858.6                                     & 9315.9                                    & 7037.4                                                  & \textbf{11161}                         & 1522.9                 & (1513.1)    \\
      James Bond        & 29.0                   & 302.8          & 366.5                                      & \textbf{517.0}                            & 462.7                                                   & 445                                    & 280.9                  & (60.9)      \\
      Kangaroo          & 52.0                   & 3035.0         & 3617.4                                     & 724.1                                     & 838.2                                                   & \textbf{4098}                          & 271.2                  & (298.0)     \\
      Krull             & 1598.0                 & 2665.5         & 3681.6                                     & 5663.3                                    & 6616.4                                                  & 7782                                   & \textbf{7813.7}        & (1598.0)    \\
      Kung Fu Master    & 258.5                  & 22736.3        & 14783.2                                    & \textbf{30944.8}                          & 21759.8                                                 & 21420                                  & 20169.8                & (8206.7)    \\
      Ms Pacman         & 307.3                  & 6951.6         & 1318.4                                     & 1281.2                                    & 999.1                                                   & 1327                                   & \textbf{1356.8}        & (775.8)     \\
      Pong              & -20.7                  & 14.6           & -5.4                                       & \textbf{20.1}                             & 14.6                                                    & 18                                     & 12.6                   & (10.0)      \\
      Private Eye       & 24.9                   & 69571.3        & 86.0                                       & 96.7                                      & 100.0                                                   & \textbf{882}                           & 405.5                  & (1144.3)    \\
      Qbert             & 163.9                  & 13455.0        & 866.3                                      & \textbf{13781.9}                          & 745.7                                                   & 3405                                   & 685.0                  & (68.3)      \\
      Road Runner       & 11.5                   & 7845.0         & 12213.1                                    & \textbf{17751.3}                          & 9614.6                                                  & 15565                                  & 8164.2                 & (4066.8)    \\
      Seaquest          & 68.4                   & 42054.7        & 558.1                                      & \textbf{1100.2}                           & 661.3                                                   & 618                                    & 476.8                  & (88.7)      \\
      Up n' Down        & 533.4                  & 11693.2        & 10859.2                                    & \textbf{17264.2}                          & 3546.2                                                  & N/A                                    & 7745.0                 & (8515.7)    \\
      \midrule
      Normalized mean   & 0.000                  & 1.000          & 0.616                                      & \textbf{1.943}                            & 1.046                                                   & 1.12                                   & 0.884                  &             \\
      Normalized median & 0.000                  & 1.000          & 0.396                                      & \textbf{1.090}                            & 0.289                                                   & 0.49                                   & 0.152                  &             \\
      \bottomrule
    \end{tabularx}}
\end{table}

\vspace*{-5mm}

\begin{table}[H]
  \centering
  \caption{Mean scores for the ablation studies.}
  \label{tab:ablations}
  \vspace{\belowcaptionskip}
  \small
  \begin{tabularx}{\linewidth}{l@{}ZZZZZ}
    \toprule
    \textbf{Ablation}           & \textbf{Boxing} & \textbf{Breakout} & \textbf{Kung Fu Master} & \textbf{Ms Pacman} & \textbf{Pong} \\
    \midrule
    \methodname{} (default)     & 83.4            & 50.7              & 20169.8                 & 1356.8             & 12.6          \\
    No augmentations        & 1.6             & 9.1               & 6031.4                  & 847.6              & -20.6         \\
    No action stacking      & 63.3            & 22.9              & 11981.2                 & 651.4              & -3.1          \\
    No frame stacking       & 12.1            & 8.2               & 17833.2                 & 708.4              & 3.2           \\
    No temporal consistency & 0.0             & 6.4               & 17008.8                 & 754.3              & -20.6         \\
    Sample-contrastive    & 77.4            & 25.1              & 19237.2                 & 1162.6             & 6.4           \\
    %
    \bottomrule
  \end{tabularx}
\end{table}

\vspace*{-5mm}

\begin{center}
  \begin{minipage}[H]{0.48\linewidth}
    \centering
    \begin{table}[H]
      \centering
      \caption{Total training times of various methods on the Atari 100k benchmark.
        They are approximated for an NVIDIA V100 GPU.}
      \label{tab:runtimes}
      \vspace{\belowcaptionskip}
      \small
      \begin{tabularx}{\linewidth}{lZ}
        \toprule
        \textbf{Method}      & \textbf{Runtime (hours)} \\
        \midrule
        SPR                  & 2.3                      \\
        \methodname{} (ours) & 3 \\
        DreamerV3            & 12                       \\
        TWM                  & 20                       \\
        EfficientZero        & 29                       \\
        IRIS                 & 168                      \\
        SimPLE               & 240                      \\
        \bottomrule
      \end{tabularx}
    \end{table}
  \end{minipage}
  \hfill
  \begin{minipage}[H]{0.48\linewidth}
    \centering
    \begin{table}[H]
      \centering
      \caption{Detailed time breakdown. Percentages in the lower half are relative to the default setting,
        obtained by enabling or disabling components.}
      \label{tab:time-details}
      \vspace{\belowcaptionskip}
      \small
      \begin{tabularx}{\linewidth}{lZ}
        \toprule
        \textbf{Component}     & \textbf{Percentage} \\
        \midrule
        Total training         & 100 \%              \\  
        World model training   & 63 \%               \\
        Policy training        & 37 \%               \\
        \midrule
        $-$ No augmentations   & $-16$ \%            \\  
        $-$ No action stacking & $-0.1$ \%           \\  
        $-$ No frame stacking  & $-0.1$ \%           \\  
        $+$ With decoder       & $+19$ \%            \\  
        \bottomrule
      \end{tabularx}
    \end{table}
  \end{minipage}
\end{center}

\clearpage

\section{Additional Ablations} \label{sec:add-ablations}

\begin{table}[H]
  \centering
  \caption{Mean scores for additional ablation studies.}
  \label{tab:add-ablations}
  \begin{tabularx}{\linewidth}{l@{}ZZZZZ}
    \toprule
    \textbf{Ablation}             & \textbf{Boxing} & \textbf{Breakout} & \textbf{Kung Fu Master} & \textbf{Ms Pacman} & \textbf{Pong} \\
    \midrule
    \methodname{}                 & 83.9            & 42.2              & 22626.2                 & 1134.0             & 14.8          \\
    Projector${} \times 4$ & 79.1            & 23.7              & 22481.4                 & 1176.1             & 6.7          \\  
    Transition${} \times 4$ & 82.9            & 27.1              & 23779.4                 & 1208.7             & 13.5          \\  
    Actor/critic${} \times 4$ & 77.9 & 29.7 & 17627.8 & 787.5 & 1.7 \\  
    Horizon $= 15$            & 57.9            & 30.9              & 21579.4                 & 978.1             & -0.4           \\  
    Training${} \times 4$      & 64.1 & 29.0 & 14035.8 & 572.4 & 13.3           \\
    Stack size $= 8$         & 68.1            & 21.8              & 19171.2                 & 1154.7             & 11.5           \\  
    Stack size $= 12$        & 35.3 & 14.4 & 16423.6 & 1042.8 & 12.6 \\
    Recurrent transition & 71.7 & 11.8 & 18639.4 & 1105.5 & 13 \\
    Recurrent predictors & 87.8 & 9.4 & 16086.2 & 1032.1 & 10.5 \\  
    \bottomrule
  \end{tabularx}
\end{table}

In this section we provide additional ablation studies to analyze the effect of
increasing the model size and training time. The results are shown in
\cref{tab:add-ablations}. We evaluated these additional ablation studies
on 5 instead of 10 random seeds. We observe that the default configuration of
\methodname{} performs best in most cases. The ablations are as follows:
\begin{enumerate}
  \item Projector${} \times 4$: We increase the hidden dimension
        of the projector network from $2048$ to $8192$. This increases the
        number of parameters of this network from $9.5$M to $88$M.
  \item Transition${} \times 4$: We increase the hidden dimension
        of the transition network from $1024$ to $4096$. This increases the
        number of parameters of this network from $5.2$M to $71.3$M.
  \item Actor/critic${} \times 4$: We increase the hidden dimension
        of the actor and critic networks from $512$ to $2048$. This increases
        the total number of parameters of the agent from $0.5$M to $8.4$M.
  \item Horizon${} = 15$: We increase the imagination horizon $H$ from $10$
        to $15$ steps, and reduce the imagination batch size to $2048$ to keep
        the effective batch size constant.
  \item Training${} \times 4$: We train the world model and the
        agent with two batches per environment step instead of one batch every
        second step. This effectively multiplies the training time by four and
        is similar to the training time of DreamerV3.
  \item Stack size${} = 8$: We stack $8$ frames and actions instead of $4$.
  \item Stack size${} = 12$: We stack $12$ frames and actions instead of $4$.
  \item Recurrent transition: We incorporate a recurrent layer
        (LSTM) into the transition network, placing it after the five hidden
        linear layers and before the final output layer. This requires several
        changes in the implementation, since the recurrent states must be
        maintained and passed between steps.
  \item Recurrent predictors: We make the transition, reward, and
        terminal distributions recurrent by introducing a shared three-layer MLP
        followed by an LSTM layer. The output of the LSTM is then fed into a
        two-layer transition head, as well as the reward and terminal networks.
\end{enumerate}


\clearpage

\section{Implementation Details} \label{sec:impl-details}

Implementing a world model involves numerous design choices, many of which may
seem arbitrary at the first glance or are obscured in the source code. In the
following, we explain all of our implementation details.

\paragraph{Stacking and preprocessing.}
As detailed in \cref{sec:ingredients}, we stack the $m$ most recent
observations and actions, with ${m = 4}$. Frame stacking also plays a role in
our representation learning approach, with observations ${\vo}$ and ${\vo}'$
sharing information from three subsequent frames. For data augmentation, we
apply the transformations proposed for Atari by \citet{drq}, i.e., random
shifts and imagewise intensity jittering.

\paragraph{Distributions.}

We model the transition distribution using independent normal distributions
with unit variance, i.e., ${\ptra(\vy' \cond \vy, \va) =
  \mathcal{N}(\vy'\cond\mu_\theta(\vy, \va), \mI_\ydim)}$, where ${\mu_\theta}$
is a neural network computing the mean vector. We chose this distribution since
the loss function reduces to minimizing the mean squared error ${\tfrac{1}{2}
      \Vert \mu_\theta(\vy, \va) - \vy' \Vert_2^2}$. Also, the mean that we use for
prediction is available without further computation. We model the reward
distribution $\prew(r \cond \vy, \va, \vy')$ using discrete regression with
two-hot encoded targets and symlog predictions, as recently proposed by
\citet{dreamerv3}. Although not yet being a widely used approach, it makes
reward prediction stable across different scales without the need for
domain-specific reward normalization or hyperparameter tuning. We model the
terminal distribution using a Bernoulli distribution, i.e., ${\pend(e \cond
  \vy, \va, \vy') = \text{Bernoulli}(e\cond\sigmoid_\theta(\vy, \va, \vy'))}$,
where ${\sigmoid_\theta}$ is a neural network computing the terminal
probability. We chose this distribution since it is a common choice for
distributions with binary support; the loss function reduces to the binary
cross-entropy, and the mode can be computed by ${[\sigmoid_\theta(\vy, \va,
          \vy') \ge 0.5]}$ with the squared brackets being Iverson brackets.

\paragraph{Architecture.}
All networks use SiLU nonlinearities \citep{gelu} to prevent dead ReLUs,
especially given that the data is coming form an ever-changing replay buffer.
Furthermore, we employ layer normalization \citep{layer-norm} in all networks.
The encoder $\fenc$ consists of four convolutional layers with a kernel size of
$4$, stride of $2$, and padding size of $1$, followed by a linear layer that
computes representations of dimension ${\ydim = 512}$. To stabilize training,
the representations are also normalized using layer normalization; refer to
\cref{sec:methodology} for our motivation. The projector network is an MLP with
two hidden layers of dimension $2048$, computing embeddings of dimension $\zdim
  = 2048$; these dimension have been proven to strike a good balance between
qualitative performance and architecture size for VICReg \citep{duality}. The
predictor network uses the same architecture as the projector network. The
network of the transition distribution is an MLP with five hidden layers of
dimension $1024$, and a residual connection from the input to the output. The
networks of the reward distribution, terminal distribution, policy, and value
function are MLPs with two hidden layers of dimension $1024$. We use the AdamW
optimizer \citep{adamw} for all networks and loss functions.

\paragraph{Actor-critic.}
We estimate advantages using generalized advantage estimation \citep{gae} and
calculate multi-step truncated $\lambda$-returns \citep{rl-intro} as the target
for the value function. To improve exploration and prevent early convergence to
suboptimal policies, we add the entropy of the policy to the objective
\citep{rl-entropy,a3c}. Additionally, we adopt the following strategies from
DreamerV3 \citep{dreamerv3}, which have demonstrated success across various
environments and reward scales without domain-specific fine-tuning. For
advantage computation, the returns are normalized by mapping the
5\textsuperscript{th} and the 95\textsuperscript{th} percentile to 0 and 1,
respectively. The value function utilizes the same discrete regression approach
as the reward predictor, i.e., two-hot encoded targets and symlog predictions.
A target network, which is the exponential moving average of the online value
network, computes additional targets for the value function. This allows for
estimating returns using the online network instead of the target network.

\clearpage



\newcommand\mycommfont[1]{\rmfamily\textcolor{gray}{#1}}
\setlength\algotitleheightrule{0.5\algotitleheightruledefault}
\setlength{\algomargin}{0pt}
\setlength{\interspacetitleruled}{3pt}
\SetAlCapHSkip{0pt}
\begin{figure}[H]
  \centering
  \begin{minipage}{.75\linewidth}
    \begin{algorithm2e}[H]
      \caption{\methodname{}'s main training procedure.}
      \label{algo:main}
      \vskip 2pt
      \SetAlgoNoLine
      \DontPrintSemicolon
      \SetCommentSty{mycommfont}
      \SetKwComment{Comment}{\# }{}
      \KwIn{environment $\mathcal{E}$, environment steps $M$, imagination horizon $H$}
      initialize replay buffer $\mathcal{D}$\;
      initialize networks of the world model and the policy\;
      \For{$i \in \{1, \ldots, M\}$}{
        execute action in $\mathcal{E}$ according to policy $\ppi$\;
        store observed transition in $\mathcal{D}$\;
        \Comment{all following computations are batch-wise}
        \If{world model update}{
          sample batch of transitions $\tau \sim \mathcal{D}$\;
          estimate $\loss_\text{Repr.}(\theta)$ and $\loss_\text{Dyn.}(\theta)$ according to \labelcref{eq:repr-loss}, \labelcref{eq:dyn-loss}\;
          update $\theta$ to minimize the losses
        }
        \If{policy update}{
          \Comment{sample from arbitrary time steps}
          sample batch of observations $\mO \sim \mathcal{D}$\;
          encode observations $\mY_1 = \fenc(\mO)$\;
          \For{$t \in \{1, \ldots, H\}$}{
            select actions $\mA_t \sim \ppi(\mA_t \cond \mY_t)$\;
            predict $\mY_{t+1} \sim \ptra(\mY_{t+1} \cond \mY_t, \mA_t)$\;
            predict $\vr_{t+1} \sim \prew(\vr_{t+1} \cond \mY_t, \mA_t, \mY_{t+1})$\;
            predict $\ve_{t+1} \sim \prew(\ve_{t+1} \cond \mY_t, \mA_t, \mY_{t+1})$\;
          }
          update $\phi$ actor-critic style using trajectories\;
        }
      }
    \end{algorithm2e}
  \end{minipage}
\end{figure}

\begin{table}[H]
  \centering
  \captionsetup{width=0.75\linewidth}
  \caption{Summary of all hyperparameters. Note that we use the original
    coefficients for VICReg.}
  \label{tab:hyperparameters}
  \vspace{\belowcaptionskip}
  \small
  \begin{tabularx}{0.75\linewidth}{lcY}
    \toprule
    \textbf{Hyperparameter}           & \textbf{Symbol} & \textbf{Value}             \\
    \midrule
    Dimensionality of $y$             & $\ydim$         & \num{512}                  \\
    Dimensionality of $z$             & $\zdim$         & \num{2048}                 \\
    Consistency coefficient           & $\eta$          & 12.5                       \\
    Covariance coefficient            & $\rho$          & 1.0                        \\
    Variance coefficient              & $\nu$           & 25.0                       \\
    Frame resolution                  & --              & ${64 \times 64}$           \\
    Grayscale frames                  & --              & No                         \\
    Terminal on loss of life          & --              & Yes                        \\
    Frame and action stacking         & $m$             & 4                          \\
    Random shifts                     & --              & 0--3 pixels                \\
    Discount factor                   & $\gamma$        & 0.997                      \\
    $\lambda$-return parameter        & $\lambda$       & 0.95                       \\
    Entropy coefficient               & --              & \num{1e-3}                 \\
    Target network decay              & --              & 0.98                       \\
    World model training interval     & --              & Every 2nd environment step \\
    Policy training interval          & --              & Every 2nd environment step \\
    Environment steps                 & $M$             & \num{100000}               \\
    Initial random steps              & --              & \num{5000}                 \\
    World model batch size            & --              & \num{1024}                 \\
    World model learning rate         & --              & \num{6e-4}                 \\
    World model warmup steps          & --              & \num{5000}                 \\
    World model weight decay          & --              & \num{1e-3}                 \\
    World model gradient clipping     & --              & 10.0                       \\
    Imagination batch size            & --              & \num{3072}                 \\
    Imagination horizon               & $H$             & 10                         \\
    Actor-critic learning rate        & --              & \num{2.4e-4}               \\
    Actor-critic gradient clipping    & --              & 100.0                      \\
    Policy temperature for evaluation & --              & 0.5 (0.01 for Freeway)     \\
    Random actions during collection  & --              & 1\%                        \\
    \bottomrule
  \end{tabularx}
\end{table}

\end{document}